\documentclass[acmlarge]{acmart}

\usepackage[utf8]{inputenc}
\usepackage[T1]{fontenc}
\usepackage{graphicx}
\usepackage{adjustbox}
\usepackage{tabularx}
\usepackage{subcaption}
\usepackage[english]{babel}
\usepackage[figuresright]{rotating}
\usepackage{makecell}
\usepackage{array}
\usepackage{caption}
\usepackage{multirow}
\usepackage{url}
\usepackage{float}
\usepackage{hyperref}
\usepackage{adjustbox}
\usepackage{threeparttable}
\usepackage{rotating} 
\usepackage{array}
\usepackage{makecell}
\usepackage{tikz}
\usepackage{afterpage}
\usepackage{longtable,lscape}
\usepackage{enumitem}
\usepackage{color, colortbl}
\usepackage{makecell}
\usepackage{wrapfig}
\usepackage[linesnumbered,ruled,vlined]{algorithm2e}
\definecolor{Gray}{gray}{0.9}
\definecolor{LightGray}{gray}{0.6}
\definecolor{RED}{RGB}{255,0,0}
\usepackage[most]{tcolorbox}
\hypersetup{colorlinks=true, citecolor={magenta}, filecolor=blue, linkcolor=black, urlcolor=blue}

\usepackage{xcolor}
\definecolor{green(munsell)}{rgb}{0.0, 0.66, 0.47}
\definecolor{cadmiumgreen}{rgb}{0.0, 0.42, 0.24}
\definecolor{cobalt}{rgb}{0.0, 0.28, 0.67}
\definecolor{amber(sae/ece)}{rgb}{1.0, 0.49, 0.0}
\newlength\MAX  

\newlength\BARSIZE  

\newcommand*\ChartBarBlue[1]{\textcolor{cobalt}{\rule{\BARSIZE}{2ex}}}
\newcommand*\ChartBarGreen[1]{\textcolor{green(munsell)}{\rule{\BARSIZE}{2ex}}}
\newcommand*\ChartBarOrange[1]{\textcolor{amber(sae/ece)}{\rule{\BARSIZE}{2ex}}}

\definecolor{Gray}{gray}{0.9}
\definecolor{Gray2}{gray}{0.8}

\newcolumntype{a}{>{\columncolor{Gray}}c}
\newcolumntype{b}{>{\columncolor{white}}c}

\AtBeginDocument{%
  \providecommand\BibTeX{{%
    \normalfont B\kern-0.5em{\scshape i\kern-0.25em b}\kern-0.8em\TeX}}}

\newcommand{\lakmal}[1]{\textcolor{black}{#1}}

\setcopyright{acmlicensed}
\acmJournal{IMWUT}
\acmYear{2024} \acmVolume{8} \acmNumber{2} \acmArticle{46} \acmMonth{5}\acmDOI{10.1145/3659591}

\begin{document}

\title{M3BAT: Unsupervised Domain Adaptation for Multimodal Mobile Sensing with Multi-Branch Adversarial Training}

\author{Lakmal Meegahapola}
\email{lakmal.meegahapola@epfl.ch}
\authornote{now at ETH Zurich, Switzerland}
\orcid{0000-0002-5275-6585}
\affiliation{
\institution{Idiap Research Institute \& EPFL} \country{Switzerland}}

\author{Hamza Hassoune}
\orcid{0009-0000-0239-9228}
\affiliation{
\institution{Idiap Research Institute \& EPFL} \country{Switzerland}}

\author{Daniel Gatica-Perez}
\orcid{0000-0001-5488-2182}
\affiliation{\institution{Idiap Research Institute \& EPFL} \country{Switzerland}}

\renewcommand{\shortauthors}{Meegahapola et al.}
\renewcommand{\shorttitle}{Unsupervised Domain Adaptation for Multimodal Mobile Sensing with Multi-Branch Adversarial Training}

\begin{abstract}

Over the years, multimodal mobile sensing has been used extensively for inferences regarding health and well-being, behavior, and context. However, a significant challenge hindering the widespread deployment of such models in real-world scenarios is the issue of distribution shift. This is the phenomenon where the distribution of data in the training set differs from the distribution of data in the real world---the deployment environment. While extensively explored in computer vision and natural language processing, and while prior research in mobile sensing briefly addresses this concern, current work primarily focuses on models dealing with a single modality of data, such as audio or accelerometer readings, and consequently, there is little research on unsupervised domain adaptation when dealing with multimodal sensor data. To address this gap, we did extensive experiments with domain adversarial neural networks (DANN) showing that they can effectively handle distribution shifts in multimodal sensor data. Moreover, we proposed a novel improvement over DANN, called \textbf{M3BAT}, unsupervised domain adaptation for \textbf{m}ultimodal \textbf{m}obile sensing with \textbf{m}ulti-\textbf{b}ranch \textbf{a}dversarial \textbf{t}raining, to account for the multimodality of sensor data during domain adaptation with multiple branches. Through extensive experiments conducted on two multimodal mobile sensing datasets, three inference tasks, and 14 source-target domain pairs, including both regression and classification, we demonstrate that our approach performs effectively on unseen domains. Compared to directly deploying a model trained in the source domain to the target domain, the model shows performance increases up to 12\% AUC (area under the receiver operating characteristics curves) on classification tasks, and up to 0.13 MAE (mean absolute error) on regression tasks. 
 
\end{abstract}



\maketitle

\section{Introduction}\label{sec:introduction}

In recent years, the prevalence of mobile and wearable devices equipped with multimodal sensors has increased significantly, offering a wide range of applications in health, well-being, context awareness, and user experience \cite{lane2010survey, meegahapola2020smartphone}. These sensors can capture diverse data, including accelerometers, gyroscopes, photoplethysmography (PPG) readings, and location, as well as device usage data like application usage and typing and touch events. This wealth of data presents exciting opportunities for understanding human behavior \cite{assi2023complex}, physiological responses \cite{gedam2021review}, and contextual information \cite{yurur2014context} in an unobtrusive manner. Some examples include activity recognition \cite{baldominos2019comparison, assi2023complex, bouton2022your}, stress detection \cite{matton2023contrastive, healey2005detecting, mishra2020evaluating}, mood inference \cite{likamwa2013moodscope, servia2017mobile, meegahapola2023generalization}, eating and drinking behavior understanding \cite{meegahapola2021examining, meegahapola2020protecting, santani2018drinksense, meegahapola2021one}, and social context recognition \cite{kammoun2023understanding, meegahapola2020alone, mader2024learning}. However, despite the growing interest in utilizing multimodal sensor data, several challenges must be addressed to fully harness their potential in deployment settings. One such under-explored challenge is the generalization of models across different users, populations, and environments \cite{xu2023globem, meegahapola2023generalization, adler2022machine}. As each individual exhibits unique behavioral patterns and physiological responses, building models that are robust and adaptable across diverse populations poses a challenge \cite{muller2021depression, schelenz2021theory}. Additionally, variations in the context of data collection can significantly impact sensor readings and behavior patterns (e.g., training a model in Italy and expecting it to work for people in India) \cite{assi2023complex, meegahapola2023generalization}. Achieving generalization is challenging due to distribution shifts in the data.

The data collected from various sensors in different environments may not align perfectly, resulting in a distribution shift between the source dataset (data that the model is trained on) and the target dataset (data that the model would encounter in deployment) \cite{chang2020systematic, varshney2022trustworthy}. These distribution shifts can have a negative impact on model performance when applied in new and unseen contexts. Addressing distribution shifts requires the use of transfer learning techniques, including robust domain adaptation approaches \cite{ganin2016domain}. Despite numerous studies exploring the applications of multimodal sensors in mobile and wearable devices, discussions around the challenges of generalization and distributional shifts have been relatively limited \cite{xu2023globem, meegahapola2023generalization}. This is in contrast to other domains, such as computer vision, natural language processing, and speech processing, where significant progress has been made in understanding and mitigating domain shifts \cite{zhou2022domain}. However, prior studies have emphasized that blindly adapting techniques from other domains to mobile sensing datasets is not trivial and needs deeper investigation because of the differences in the way data are collected, processed, and made sense of \cite{chang2020systematic, wu2023udama, xu2023globem}. Therefore, more investigations are needed in multimodal sensing settings to overcome challenges regarding distribution shifts.

In mobile sensing settings, training models often rely on large-scale datasets collected from multiple users. In deployment, models need to personalize for better performance, and having ground truth labels from users is a primary way to do this. However, obtaining labeled ground truth from users poses challenges due to the sparse nature of data collection and difficulties in acquiring accurate and reliable self-reports \cite{xu2021leveraging}. Consequently, the lack of labeled data impedes the personalization of models for individual users, making it difficult to cater to their unique characteristics and preferences. Therefore, the crucial step of adapting models to target populations (i.e., genders, age groups, countries, sub-populations, etc.) becomes essential even before personalization \cite{meegahapola2023generalization, assi2023complex}. By adapting the models to the target population, we can ensure their effectiveness in diverse contexts, providing a strong foundation for subsequent personalization efforts. Unsupervised domain adaptation (UDA) \cite{ganin2015unsupervised} techniques play a vital role in bridging the gap between different domains, rendering the models more versatile and adaptable to various users and environments. However, even though UDA has been explored in very few prior studies in mobile sensing \cite{chang2020systematic, mathur2019unsupervised, meegahapola2023generalization}, how such techniques perform when multimodal data are present has rarely been explored.

Considering these aspects, in this paper, we first conduct a statistical analysis of datasets to understand the dynamics of distribution shifts across source-target domain pairs and different sensing modalities. Then, we evaluate unsupervised domain adaptation with domain adversarial training (DANN) \cite{ganin2016domain, chang2020systematic}, and also other baselines such as maximum mean discrepancy (MMD) \cite{chang2020systematic} and adversarial discriminative domain adaptation (ADDA) \cite{tzeng2017adversarial}, on two different multimodal mobile and wearable sensing datasets across both regression and classification tasks. Then, we propose a novel model architecture for \textbf{M}ultimodal \textbf{M}obile Sensing data called \textbf{M}ulti-\textbf{B}ranch domain \textbf{A}dversarial \textbf{T}raining (\textbf{M3BAT}), showing improved performance over baselines across a majority of inference tasks. In doing this, we answer the following research questions:

\begin{itemize}[wide, labelwidth=!, labelindent=0pt]
    \item[\textbf{RQ1:}] What dynamics regarding distribution shift can be observed by conducting a statistical analysis on multimodal sensing datasets? 
    \item[\textbf{RQ2:}] Does DANN on multimodal sensing datasets lead to improved UDA performance? How does it compare to transfer learning-based fine-tuning when labels are available in the target domain? 
    \item[\textbf{RQ3:}] Does having multiple branches for different feature sets (based on modality or distribution shift-based feature groups) lead to improved UDA performance? 
\end{itemize}{}

By addressing the above research questions, this paper provides the following contributions: 
\begin{itemize}[wide, labelwidth=!, labelindent=0pt]
    \item[\textbf{Contribution 1:}] We conducted an analysis of two multimodal sensing datasets, namely WENET and WEEE. These datasets provide valuable insights into distribution shifts across various dimensions---WENET explores distribution shifts across different countries with modalities such as wifi, steps, proximity, location, screen events, app usage, activity, etc., while WEEE captures shifts across devices worn on distinct body positions with accelerometer, photoplethysmography (PPG), and gyroscope data. Our approach involved calculating Cohen's-d values for individual features and then aggregating them to discern patterns at the modality and feature set level. This analysis allowed us to pinpoint the modalities that exhibited high distribution shifts across various source and target domain pairs. For instance, in the WENET dataset, activity and screen event data demonstrated minimal difference between Italy and India, while wifi and step count features displayed substantial dissimilarity, attributable to low and high shifts, respectively. These trends contrasted across source-target pairs, underscoring the importance of a multimodality-aware architecture that accounts for individual modality and feature set level shifts during the domain adaptation process. This leads to the question of whether it is worth exploring model architectures that explicitly cater multimodality of data in unsupervised domain adaptation.
    
    \item[\textbf{Contribution 2:}] The datasets employed in our study provide a platform for exploring diverse inferences, encompassing mood, social context, and energy expenditure estimation via classification and regression tasks. In order to comprehensively assess the impact of multimodality, we transformed the datasets into tabular formats and conducted domain adaptation using domain adversarial training with gradient reversal, employing the DANN approach. Notably, our results showed an improvement in performance, demonstrating an increase of up to 8\% in AUC for classification tasks, and a reduction of 0.08 in MAE for regression tasks when compared to deploying the model directly on target domains. Remarkably, in the context of the WENET dataset, unsupervised domain adaptation demonstrated competitive performance with transfer learning-based fine-tuning, highlighting its potential to enhance performance even when not explicitly tailored to multimodality. However, for the WEEE dataset, while domain adversarial training led to performance improvement, it fell short of transfer learning. This disparity could be attributed to the presence of high-quality gold standard labels in both source and target domains in WEEE (as opposed to both subjective and objective, but silver-standard labels in WENET), which were effectively harnessed for model fine-tuning when labels were accessible in the target domain. This analysis underscores the significance of label quality and its interplay with domain adaptation techniques, offering insights into the diverse impacts of datasets and label types on overall performance.

    \item[\textbf{Contribution 3:}] We introduce an improvement to DANN, in the form of a novel architecture for domain adversarial training, denoted as \textbf{M3BAT}, which employs a multi-branch neural network structure featuring multiple encoders tailored to handle different feature sets. Each encoder is designed to accommodate specific features, taking into account factors such as the extent of distribution shifts; and features stemming from various modalities. Through the concatenation of encoder outputs, our architecture incorporates domain adversarial training techniques, including parameter annealing and staged training. Our analysis suggests that employing three branches yields more stable training for the datasets and tasks under consideration for WENET. These branches correspond to high, moderate, and low shift features. For WEEE, we employ 2-3 branches in different setups. On average, we observe an increase of up to 12\% in AUC for classification tasks, along with a reduction of up to 0.13 in MAE for regression tasks, as compared to deploying models from the source to the target domain. These findings underscore the potential advantages of our methodology in managing distribution shifts in multimodal data.
    
\end{itemize}{}

The study is organized as follows. In Section~\ref{sec:related_work}, we describe the background and related work. Then we describe the proposed architecture in Section~\ref{sec:architecture}. Section~\ref{sec:datasets} provides a description of the data used. In Section~\ref{sec:rq1}, Section~\ref{sec:rq2}, and Section~\ref{sec:rq3}, we define the methods and present results to answer RQ1, RQ2, and RQ3, respectively. We discuss implications, limitations, and future work in Section~\ref{sec:discussion}, and conclude the paper in Section~\ref{sec:conclusion}.

\section{Background and Related Work}\label{sec:related_work}

\subsection{Distribution Shift}

In the context of machine learning, distribution shift refers to the mismatch between the probability distributions of the data in the source domain (where the model is trained) and the target domain (where the model is deployed) \cite{varshney2022trustworthy}: $ p_{X,Y}(\text{source})(x,y) \neq p_{X,Y}(\text{target})(x,y) $. When this mismatch occurs as a result of epistemic uncertainty \cite{hullermeier2021aleatoric}, the model's performance can degrade significantly in the target domain, as it has not seen data from that domain during training. The epistemic uncertainty could be due to many sampling biases such as temporal bias, population bias, and social bias \cite{olteanu2019social}. Hence, in other terms, distribution shifts can arise due to various factors, such as differences in data collection settings, user preferences, environmental conditions, and cultural variations. While there are many nitty-gritty details, three primary types of distribution shift can be identified \cite{varshney2022trustworthy}: covariate shift, label shift, and concept drift. Understanding these types is crucial for effectively addressing the challenges posed by distribution shifts. The first type, namely covariate shift, also known as input shift or feature shift, occurs when the input data's distribution differs between the source and target domains, but the conditional distribution of the labels given the input remains the same. In other words, the relationship between the input features and the labels is consistent across domains, but the frequency of different feature values may vary. This can be represented as: $
p_X(\text{source})(x) \neq p_X(\text{target})(x) \quad \text{and} \quad p_{Y\mid X}(\text{source})(y\mid x) = p_{Y\mid X}(\text{target})(y\mid x) $.
Therefore, differences in data collection methods, sensor characteristics, or user behavior across different domains can cause covariate shifts. To illustrate covariate shift, consider a sentiment analysis model trained on movie reviews from the source domain (e.g., American movies) and deployed in the target domain (e.g., Indian movies). The language and writing style of the reviews may differ between the two domains, even though the sentiment expressed by the reviews is the same. In this case, the covariate shift arises from variations in language usage while the sentiment remains consistent. The second type of distribution shift, prior probability shift, also known as label shift, occurs when the label distributions are different between the source and target domains, while the conditional distributions of features given the labels are the same. It can be represented as: $ p_Y(\text{source})(y) \neq p_Y(\text{target})(y) \quad \text{and} \quad p_{X\mid Y}(\text{source})(x\mid y) = p_{X\mid Y}(\text{target})(x\mid y)$. Label shift can arise when the labeling process is biased or when the target domain has different class distributions compared to the source domain. Continuing with the sentiment analysis example, label shift may occur if the sentiment expression in movie reviews is perceived differently in different cultures. For instance, positive reviews in the source domain might be labeled as negative in the target domain due to cultural differences in how sentiments are conveyed. Finally, the third type of distribution shift, concept drift, is a much more complex aspect to mitigate \cite{varshney2022trustworthy} and not the focus of this paper. In this paper, the primary objective is to handle covariate shifts. 

\subsection{Unsupervised Domain Adaptation}

Unsupervised domain adaptation (UDA) is a transfer learning technique used to mitigate the effects of distribution shifts between the source and target domains without requiring labeled data from the target domain \cite{varshney2022trustworthy, ganin2015unsupervised, chang2020systematic}. The process was primarily developed to handle the covariate shift. As described in \cite{he2023domain}, there are multiple methods that achieve this. Self-supervised domain adaptation, mostly explored in computer vision \cite{xu2019self, achituve2021self, ma2022self}, is a common technique that learns domain invariant features through pretext learning tasks where the target can be obtained without supervision. Statistical divergence can also be used in this context. The approach seeks to derive features that are consistent across domains by minimizing domain discrepancies in a latent feature space. Different techniques are used to implement this: Maximum mean discrepancy (MMD) \cite{rozantsev2018beyond}, Correlation alignment (CORAL) \cite{sun2016deep}, or contrastive domain discrepancy (CDD) \cite{kang2019contrastive}. Furthermore, Test-Time adaptation (TTA) \cite{liang2023comprehensive, liang2020we, wang2020tent, iwasawa2021test} can also mitigate cases where the test data distribution differs from the training data. TTA relies on adapting a pre-trained model from source domain to the target domain before making predictions, not requiring to finetune the model on labeled data from the target distribution and not even requiring the source dataset.
Lastly, domain adversarial training (DANN), introduced by Ganin et al. \cite{ganin2016domain}, is a popular approach for UDA \cite{wang2018unsupervised, zhang2018collaborative, ganin2016domain}. Variations of this technique have shown to perform well on sensor data \cite{wu2023udama, chang2020systematic}, motivating us to explore the technique further. While DANN can indirectly influence the alignment of label distributions or prior probabilities between domains through the shared feature space, it is not the primary mechanism for addressing prior probability shifts. Hence, this process is suited to cover both covariate and label shifts to varying extents. The key idea is to learn a feature representation that is domain-invariant, enabling the model to generalize well across domains. In domain adversarial training, a domain discriminator is introduced along with the primary task model (e.g., classification or regression). The domain discriminator aims to predict the domain of the input data (source or target) based on the feature representation learned by the primary task model. Simultaneously, the primary task model tries to minimize the task-specific loss and maximize the domain discriminator’s confusion, effectively aligning the feature distributions between the source and target domains. To increase the confusion, gradient reversal can be used by multiplying the loss by $-\lambda$ ($\lambda$ is a scaler) when propagating loss to the feature extractor. The domain discriminator, in turn, tries to distinguish between the source and target domains accurately. This adversarial process encourages the primary task model to learn features that are less sensitive to domain variations and more transferable between domains, leading to improved generalization in the target domain. Consequently, unsupervised domain adaptation with domain adversarial training provides a powerful solution to adapt models to new domains and improve their performance in diverse real-world settings, such as multimodal mobile and wearable sensing.

\subsection{Adapting Models to Individuals}

Several studies have delved into domain adaptation by treating individuals as distinct domains, focusing on training models with data from many individuals and adapting them to an unseen individual \cite{gong2023dapper}. This approach, while highly valuable, presents a relatively less noisy and more straightforward domain adaptation task similar to personalizing the model for the target individual, but without having labeled data. Data used in the adaptation process comes only from a certain individual. In our work, however, we confront a more nuanced scenario. Our domain adaptation tasks involve adapting the model across larger source and target domains, broader than individuals. This undertaking adds layers of complexity, as it requires addressing the challenges posed by variations across geographic regions, body positions, and individual behaviors within the target domain. As prior work has highlighted \cite{meegahapola2023generalization, assi2023complex}, it is worth adapting models to larger contexts before personalizing models because we might not be able to access data from individuals, but we might have access to data from the broader target domain an individual belongs to. This is the aim of this paper. Hence, we introduce novel considerations and methodologies to tackle this more challenging and multifaceted adaptation task.

\subsection{Mobile Sensing for Inferences Regarding Health and Well-Being, Behavior, and Context}

Mobile sensing using smartphones and wearable devices has facilitated the development of context-aware systems that can infer various aspects related to health and well-being, behavior, and context \cite{lane2010survey}. These studies leverage diverse sensor data captured by mobile devices, including accelerometer, gyroscope, gps, heart rate monitor, proximity, bluetooth, and app usage, among others, to gain insights into individuals' activities, mood, social context, and energy expenditure \cite{meegahapola2020smartphone, bangamuarachchi2023inferring, lane2010survey, amarasinghe2023multimodal, bangamuarachchi2022sensing}. Various studies have explored mood inference, aiming to understand and predict users' emotional states \cite{likamwa2013moodscope, servia2017mobile}. Servia-Rodríguez et al. \cite{servia2017mobile} collected a large-scale dataset from multiple countries to infer binary mood using the circumplex mood model with population-level models. Mood instability has also been examined using mood reports and phone sensor data \cite{morshed2019prediction, zhang2019inferring}. Context-aware systems have been extended to infer social context, including whether individuals are alone or with others during different activities \cite{meegahapola2020alone, meegahapola2021examining}. For example, Meegahapola et al. \cite{meegahapola2020alone} used sensor data from Switzerland and Mexico to infer social context during eating activities, while another study \cite{meegahapola2021examining} examined the social context of young adults during alcohol drinking episodes. Further, energy expenditure estimation (EEE) plays a crucial role in understanding and managing chronic diseases like obesity, diabetes, and metabolic disorders \cite{ee_survey_paper, amarasinghe2023multimodal}. It also enables personalized health management by providing insights into physical activity, energy consumption, and net calorie intake \cite{lagerros}. Wearable devices such as fitness trackers and smartwatches have been widely used for EEE due to their convenience and capability to measure activity, heart rate, and sleep patterns \cite{health_wearable_devices}. These devices overcome the limitations of costly gold standard EEE methods and have been positioned at various body locations to estimate energy expenditure \cite{cvetkovic, rothney}. Overall, the existing literature in the field of multimodal sensing offers valuable insights and tools for inferring various attributes from smartphone and wearable sensor data. However, there is a research gap in understanding generalization and distributional shifts across many different settings, which this paper aims to address in the context of UDA. The proposed UDA approach seeks to improve the generalization of inference models, making them more adaptable and robust in diverse settings.

\subsection{Domain Generalization in Mobile Sensing}

Achieving model generalization across multiple domains or datasets has been a challenging problem in the machine learning community. Transfer learning addresses this issue through domain adaptation, where the model can access some data with labels from the target domains in addition to the source domains \cite{varshney2022trustworthy}. A more challenging task is domain generalization, where the model can only access data from the source domains \cite{xu2023globem}. In mobile sensing settings, Xu et al. \cite{xu2023globem} examined this problem and suggested a model based on multi-task neural networks to create a robust model that would work well in target domains without access to data or labels when training. Even though the performance increase that they reported was not high (2\% to 5\%), it was justifiable given that the inference they performed regarding depression detection is already a challenging one. In addition, they highlight that deep learning-based domain generalization techniques designed for computer vision tasks do not work well on longitudinal and multimodal passive sensing data. While they also used multimodal data, they focused on domain generalization and not unsupervised domain adaptation, which is a different problem setting. Another study by Qian et al. \cite{qian2021latent} introduced the Generalizable Independent Latent Excitation (GILE) method for domain generalization. GILE is highly adaptable to varying activity patterns among individuals. It automatically disentangles domain-agnostic and domain-specific features, minimizing correlations between them. GILE's end-to-end training with three loss functions enhances its expressiveness and informativeness, and the empirical results on benchmark datasets demonstrate its superiority in handling domain shifts and improving model generalization across individuals. While the above papers looked into domain generalization, our analysis suggests a shift from domain generalization to domain adaptation, allowing the model to access a small fraction of unlabelled data from target domains.

\subsection{Domain Adaptation in Mobile Sensing}

A common approach for conducting domain adaptation is using different loss functions. Rey et al. \cite{fortes2022learning} used contrastive loss, while Chang and Mathur et al. \cite{chang2020systematic} used maximum mean discrepancy (MMD) loss. Many other studies \cite{ozyurt2022contrastive, sanabria2021contrasgan, cao2023pirl} also used contrastive learning-based techniques and looked into aligning features of the latent space to achieve domain adaptation. In this paper, we opted to leverage adversarial training over contrastive learning-based or other discrepancy-based techniques for several reasons. Firstly, our domain adaptation task involved large distribution shifts between source and target domains, where data distributions differed substantially. Domain adversarial training is suited to address and mitigate such domain discrepancies explicitly, making it a more appropriate choice for our scenario. Additionally, domain adversarial training has demonstrated its effectiveness in various domain adaptation tasks, offering a robust and established framework even for sensor data \cite{wu2023udama, chang2020systematic}. While contrastive learning is valuable in many machine learning applications, it may require more adaptation and nuanced design to handle domain adaptation tasks effectively in multimodal sensing setups. Given these considerations, we found domain adversarial training to be a pragmatic and reliable choice to enhance our model's performance in the context of multimodal data. 

Finally, it is also worth mentioning the UDA has been previously tried for mobile sensor data by Chang and Mathur et al. \cite{chang2020systematic}, with domain adversarial training. They also used adversarial domain adaptation, similar to ours. However, they only considered a single modality of data in their experiments, hence making the task simpler compared to our experiments, which consider features from multimodal data with varying degrees of shifts across source and target domains. For example, in single modality settings, if the data are accelerometer or audio data, depending on the shift for the specific modality, UDA techniques would facilitate adaptation. However, in mobile sensing, multiple modalities of data are present. The multimodal setting has been studied in a recent study \cite{wu2023udama}, by using all features as input using a single encoder. They also performed domain adversarial training and obtained promising results. Nevertheless, existing works have largely overlooked the potential of leveraging the multimodality inherent in data during the domain adaptation process. Consequently, techniques specifically designed to harness this multimodality in UDA for mobile sensing applications are scarce. Our paper attempts to bridge this research gap by introducing and evaluating methods that capitalize on mobile sensing data's rich and multimodal nature to enhance domain adaptation outcomes.

\vspace{0.1 in}

\subsection{Summary}

In summary, our work stands out from previous studies by specifically targeting passive sensing datasets collected from mobile and wearable devices, diverging from the commonly explored data types like images or audio \cite{pillai2024investigating, mathur2019unsupervised}. Moreover, we focus on domain adaptation instead of domain generalization \cite{xu2023globem}. Furthermore, our approach focuses on the complexity of multimodal data, as opposed to the single modality focus from most existing literature \cite{chang2020systematic} or not considering multimodality \cite{wu2023udama}. This multimodal perspective is critical given the distinct nature of mobile sensing data, which encompasses a variety of user behaviors, environmental contexts, and health indicators. With M3BAT, we implement a multi-branch approach to address the inherent challenges of multimodal mobile sensing. This framework features multiple branches, each tailored to handle distinct sets or modalities of features, thereby enhancing the model's ability to adapt. By separating the data into different streams, we hypothesize that our model can specialize in processing each modality or feature set, applying specific transformations and learning domain-invariant representations more effectively. This multi-branch strategy not only simplifies the learning process for each data type but also leads to a more refined and robust adaptation performance. Our framework allows for the nuanced integration of the diverse and heterogeneous data sources that characterize multimodal mobile sensing, such as accelerometer readings, GPS locations, activity patterns, and app usage logs. By leveraging this multi-branch architecture, we show that our model achieves superior generalization across different users and environments.
\vspace{0.1 in}
\section{M3BAT Architecture}\label{sec:architecture}

In this section, we aim to define the proposed architecture, including the intuition behind it. We will first define an unsupervised domain adaptation setting for classification and regression (Section~\ref{subsec:uda_classification}). This can be represented in a generic form similar to DANN \cite{ganin2016domain}, as shown in Figure~\ref{fig:base}. Then in Section~\ref{subsec:multibranch}, we define how multiple branches could be used in both classification and regression instead of a single encoder that outputs a feature embedding. This is also shown in Figure~\ref{fig:archi1}. Finally, in Section~\ref{subsec:training_process}, we describe how different $\lambda$ could be used for different branches, depending on the shift of input features to that branch in source and target domains, to improve the performance of the model. This is summarized in Figure~\ref{fig:archi2}.

\subsection{Unsupervised Domain Adaptation with Domain Adversarial Training (DANN)}\label{subsec:uda_classification}

Given two domains, a source domain $\mathcal{D}_s = \{ (x_i^s, y_i^s) \}_{i=1}^{n_s}$ and a target domain $\mathcal{D}_t = \{ x_j^t \}_{j=1}^{n_t}$, where $x_i^s$ and $x_j^t$ represent the input feature vectors from the source and target domains, respectively, and $y_i^s$ represents the corresponding class labels in the source domain. The goal is to learn a classifier or regressor $f(x)$ that can accurately infer targets $y$ in the target domain using the labeled source domain data and the unlabeled target domain data. The domain adversarial training process consists of three main components:

\begin{itemize}
    \item Encoder: A multi-layer perceptron neural network represented by $G(x)$, which maps the input feature vectors in dimensionally reduced shared feature space, where $G_s = G(x_i^s)$ and $G_t = G(x_j^t)$ represent the features of the source and target domain samples, respectively.

    \item Target Classifier or Regressor: A head represented by $C(G_s)$, which takes the shared features $G_s$ as input and predicts $y^s$ in the source domain.

    \item Domain Classifier: A domain discriminator represented by $D(G_s)$ and $D(G_t)$, which takes the shared features $G_s$ and $G_t$ as input, respectively, and predicts whether the features are from the source or target domain.
\end{itemize}

The overall objective function for unsupervised domain adaptation with domain adversarial training for classification or regression can be written as:

\[
\min_{G, C} \max_D \frac{1}{n_s} \sum_{i=1}^{n_s} \mathcal{L}_y(C(G(x_i^s)), y_i^s) - \frac{\lambda}{n_s+n_t} \sum_{i=1}^{n_s+n_t} \mathcal{L}_{\text{d}}(D(G(x_i^s)), D(G(x_i^t)))
\]

\noindent where $\mathcal{L}_y$ is the classification loss (e.g., cross-entropy loss) or the regression loss (e.g., mean squared error) function for the source domain samples; $\mathcal{L}_{\text{d}}$ is the adversarial loss function, such as the binary cross-entropy loss, for the domain discriminator to distinguish between the source and target domain features; $\lambda$ is a parameter that controls the trade-off between the classification/regression and adversarial loss---also known as gradient reversal layer (usually $0 \le \lambda \le 1$), The first term aims to minimize the classification loss for the source domain samples, encouraging the model to infer the targets in the source domain accurately; and the second term aims to maximize the domain discriminator's confusion between the source and target domain features, effectively aligning the feature distributions of the two domains in the shared feature space.

In both classification and regression settings, unsupervised domain adaptation with domain adversarial training is a powerful technique to adapt models trained on a labeled source domain to perform well on a different, unlabeled target domain. The adversarial training process encourages the model to learn domain-invariant features, thereby improving the model's generalization to new, unseen data from the target domain. In addition, when defining, whether to use $-\lambda$ or $+\lambda$ depends on the loss function used for the domain discriminator. When the domain discriminator is binary cross-entropy which provides a negative value, using $-\lambda$ as above works \cite{ganin2016domain}.

\subsection{Multiple Branches to Process Multimodal Data}\label{subsec:multibranch}

\begin{figure}
    \centering
    
    \begin{minipage}{0.32\textwidth}
        \centering
        \includegraphics[width=\linewidth]{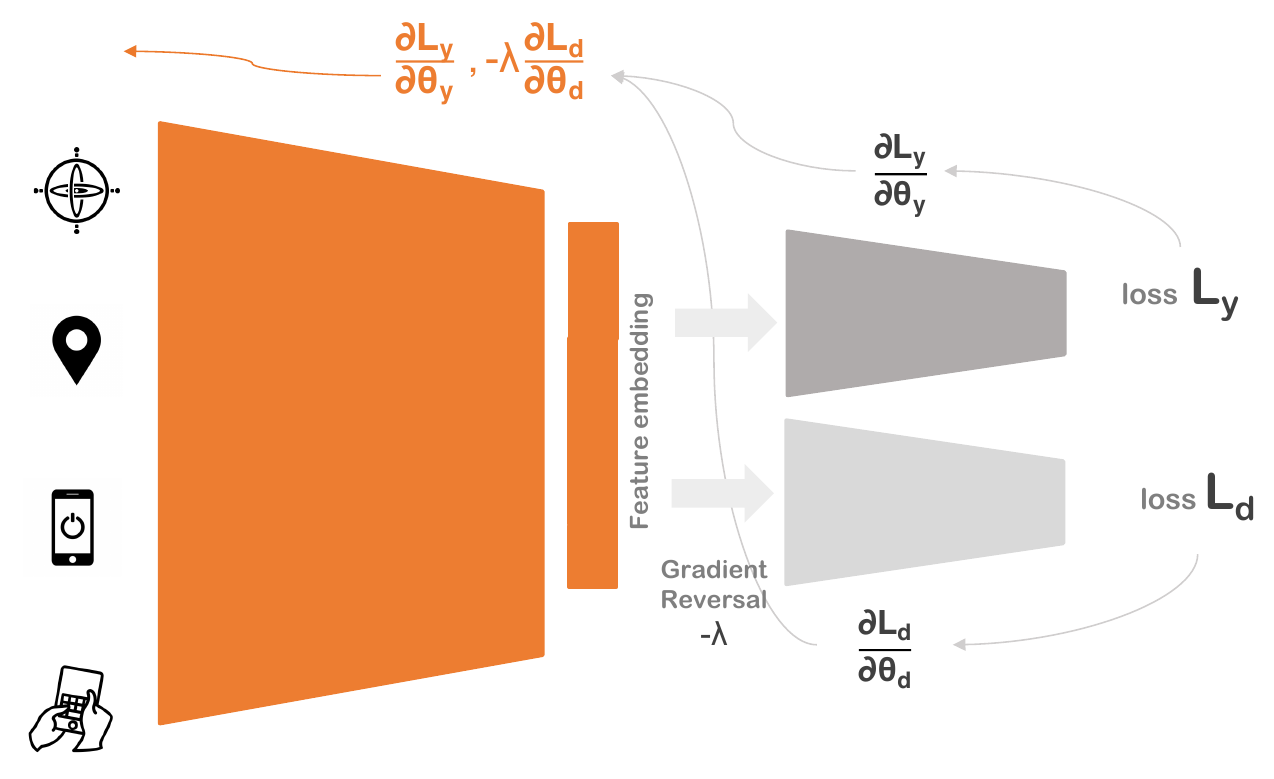}
        \caption{Base architecture for UDA with features from multimodal sensors, encoder, domain and target classifier/regressor, and gradient reversal layer.}
        \label{fig:base}
    \end{minipage}
    \hfill
    \begin{minipage}{0.32\textwidth}
        \centering
        \includegraphics[width=\linewidth]{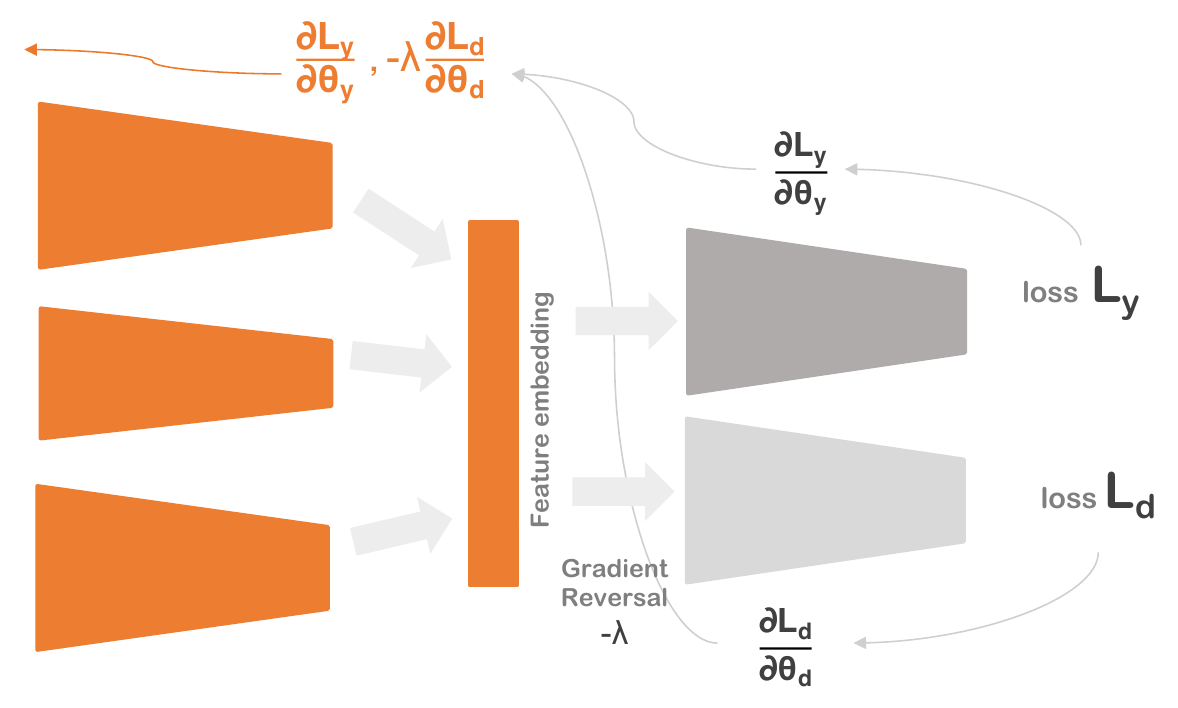}
        \caption{Modification to the base architecture to have multiple branches that concatenate to create a feature embedding.}
        \label{fig:archi1}
    \end{minipage}
    \hfill
    \begin{minipage}{0.32\textwidth}
        \centering
        \includegraphics[width=\linewidth]{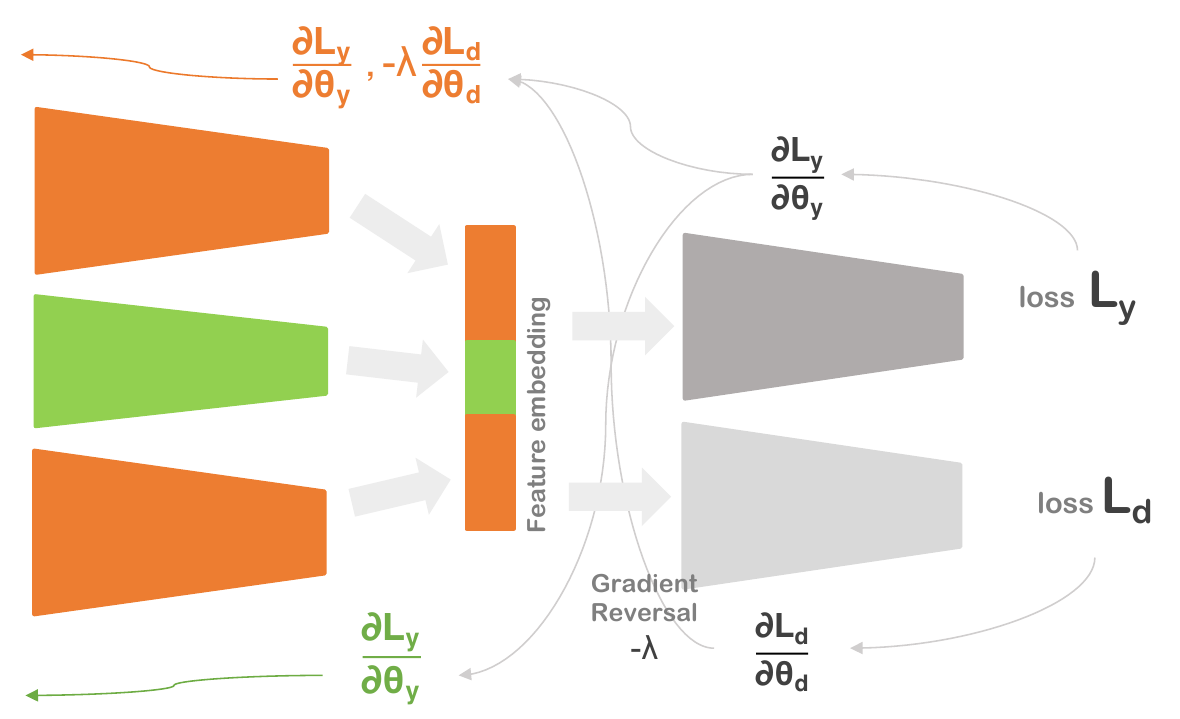}
        \caption{Using different $\lambda$ for branches depending on the average distribution shift of features in the branch. When there is little to no shift, $\lambda$$\approx$0 (green).}
        \label{fig:archi2}
    \end{minipage}

\end{figure}

To represent the setup with multiple branches for processing modalities or feature sets from multiple modalities, we can introduce separate branches. Let's denote the encoders as $E^{(m)}(x^{(m)})$, where $m$ represents the number of branches and $x^{(m)}$ is the input data from branch $m$, which could be multiple features in the tabular datasets that we consider. Each encoder processes the input data from its corresponding modality and maps it to a shared feature space. In this case, the overall feature extractor $G(x)$, which used to be a single encoder in the previous setups mentioned in Section~\ref{subsec:uda_classification}, can now be defined as a combination of these multiple encoders for each modality. We can represent this as: $
G(x) = \text{concat} \left( E^{(1)}(x^{(1)}), E^{(2)}(x^{(2)}), \ldots, E^{(M)}(x^{(M)}) \right)$. Here, $G(x)$ contains the outputs from all the encoders corresponding to the different branches, and these outputs are concatenated to form a shared feature representation that captures information from all features. With the multiple branches, the objective function for unsupervised domain adaptation with domain adversarial training can be extended to include all the modalities. With this setup, each specific encoder learns a feature representation specific to its input data, and the shared feature space created by combining the outputs of these encoders captures information from all the modalities. This approach allows the model to adapt to multiple data modalities simultaneously and improves the domain adaptation performance by considering the shared information among different modalities.

\subsection{Training Process with Multimodal Domain Adversarial Training}\label{subsec:training_process}

The training process for multimodal domain adversarial training involves a staged approach to adapt the model to the target domain while considering the distribution shift across different modalities. The process includes the following steps:

\subsubsection{Step 1: Train Common Encoder with Target Discriminator}
In the first step, we train a common encoder, $G(x)$, with only the target discriminator (classifier or regressor), $D(G(x_i^s))$. The target discriminator is responsible for performing either classification or regression. During this step, only the source domain data is used for training. The objective function for this step is to minimize the target discrimination loss, depending on whether it is regression or classification (Section~\ref{subsec:uda_classification}).

\subsubsection{Step 2: Introduce Unlabeled Target Domain Data}
After training the common encoder with the target discriminator, we introduce the unlabeled target domain data, $\mathcal{D}_t = \{ x_j^t \}_{j=1}^{n_t}$, into the training process together with domain discriminator. This is done to further align the feature distributions of the source and target domains in the shared feature space. During this step, the objective function changes a bit as we aim to perform both domain and target inferences, similar to what would happen if we used a multi-task neural network. In terms of gradient reversal, the $\lambda$=0 at this stage.

\subsubsection{Step 3a: Increase $\lambda$ for Adversarial Objective with Annealing}

To increase the impact of the adversarial objective gradually, we anneal the value of $\lambda$ from zero to one during training, as suggested in prior work \cite{yang2020curriculum, ganin2015unsupervised} ($\lambda_p = \frac{2}{1 + \exp(-\gamma \cdot p)} - 1$, where $\gamma$=10 and 0<=$p$<=1 based on the epoch). The parameter $\lambda$ controls the trade-off between the target loss and the domain loss. Increasing $\lambda$ over time encourages the common encoder to learn more domain-invariant representations, in a stable way. In Figure~\ref{fig:base}, we show the architecture at this stage, which is similar to the DANN architecture \cite{ganin2016domain}. Annealing $\lambda$ from zero to one works because it facilitates a controlled and adaptive process of aligning feature representations between the source and target domains, ultimately leading to improved domain adaptation performance. The rationale behind this approach is to start with minimal domain alignment ($\lambda$=0), allowing the model to initially focus on learning source domain knowledge without being influenced by the target domain. As training progresses and the $\lambda$ parameter gradually increases, the model is encouraged to align the feature representations of both domains \cite{ganin2015unsupervised}.

\subsubsection{Step 3b: Replace Encoder with Multiple Branches}
After training with the common encoder and gradually increasing the adversarial objective, we replace the common encoder, $G(x)$, with a multi-branch setup (Section~\ref{sec:architecture}). Each encoder, $E^{(m)}(x^{(m)})$, processes the input data from its corresponding modality or a set of features and maps it to the shared feature space. The combined feature representation is then formed by concatenating the outputs of all the branches. With this setup, steps 1 and step 2 can be followed to train the model with staging and annealing. In our experiments, we observed that having a staged process was useful for stable training. Moreover, the decision to train the target discriminator alongside the common encoder was made to enhance the stability of the training process. Starting with a common encoder simplifies the initial training stages while also ensuring satisfactory performance on the target inference. Upon transitioning to the multi-branch setup, we were presented with two options: we could either freeze the target discriminator, allowing the encoders to independently adjust to domain shifts for optimal target inference, or we could allow the target discriminator to continue fine-tuning in tandem with the new multi-branch configuration. Our preliminary experiments indicated that maintaining the adaptability of the target discriminator yielded better results, which is why this approach was adopted in our study. Additionally, we observed that commencing the training with multiple encoders from the outset did not lead to convergence, suggesting that a gradual buildup to the multi-branch system was necessary for effective training. Moreover, here maximum $\lambda=1$ across all branches. In Figure~\ref{fig:archi1}, we show the architecture at this stage.

\subsubsection{Step 3c: Increase $\lambda$ for Different Branches with Annealing}\label{subsubsec:cohensd_to_lambda}
We train the model as in Step 3b. Then, we adaptively decrease the value of $\lambda$ from 1 if needed, for each branch with annealing, until it reaches $\lambda_m$ ($0 \le \lambda_m \le \lambda$ = 1). The $\lambda_m$ values, which control the impact of the adversarial objective for each specific encoder, are determined based on the average Cohen's-d value \cite{cohen1988statistical} for each feature group. Then, the Cohen's-d values across the branches were normalized to a value between 0 and 1. If the Cohen's-d of the lowest branch is above 0.2 (above small effect size), a zero was introduced artificially before normalizing to ensure a considerable shift does not go unnoticed when performing adversarial training with different $\lambda_m$. As an example, if the Cohen's-d values were 0.8, 0.6, and 0.05, the $\lambda_m$ values would be 1, 0.58, and 0. If the Cohen's-d values from branches were 0.9 and 0.4, we would introduce a 0 to make it 0.9, 0.4, and 0 because 0.4 is above small effect size, and obtain $\lambda_m$s 1 and 0.44 for the two branches. Hence, this accounts for the distribution shift between the source and target domains specific to input features to each branch. Figure~\ref{fig:archi2}, we show the architecture at this stage. It is also worth noting that we considered two primary setups when selecting features for different branches in the encoder section of the model: \textit{Setup 1} utilized branches based on modalities in the WENET dataset, creating three branches corresponding to the modalities with the highest, lowest, and intermediate levels of shift. Due to optimization challenges discussed in Section~\ref{sec:rq2}, we limited the number of branches to three, normalizing the Cohen's-d values for these branches, with the highest shift set to 1 and the lowest to 0. Experiments were first conducted with a uniform weight ($\lambda=1$) across branches, followed by varying weights ($\lambda_m$) in accordance with the level of shift. A similar two-branch approach was adopted in the WEEE dataset, with only two modalities. This was mainly because there were only three modalities in this datasets. For \textit{Setup 2}, applied to both datasets, we sorted features by shift magnitude regardless of the modality, dividing them into three equal groups (top 33\%, bottom 33\%, and middle 33\%). Different $\lambda_m$ values were assigned to these branches based on their respective shifts. Both setups, involving tailored $\lambda_m$ adjustments, represent our methodical approach to effectively manage domain shifts in multimodal mobile sensing data. More details regarding these two set ups will be discussed in Section~\ref{subsec:rq3_methods}.

With this staged training process, the multimodal domain adversarial training algorithm can effectively adapt the model to the target domain while considering the distribution shift across different modalities or feature groups (e.g., regardless of the modality, features with a high, moderate, and low distribution shift in separate branches). The hypothesis is that this approach would allow the model to learn domain-invariant representations capable of capturing relevant information from all modalities, improving the generalization and adaptation performance to new, unseen data in the target domain.

\section{Datasets and Inferences}\label{sec:datasets}

To examine our architecture, we used multiple datasets. Both these datasets have been used in previous publications, and inferences that can be made with them too, are defined. Hence, the objective is to perform the same inferences while examining the proposed architectures. 

\subsection{WENET: Multimodal Smartphone Sensing Dataset from 8 Countries}

The WENET dataset comes from our previous work \cite{meegahapola2023generalization}. The data collection spanned four weeks and involved over 670 college students from eight universities in eight countries: Italy, Denmark, the United Kingdom, China, India, Mongolia, Paraguay, and Mexico. Participants contributed three distinct types of data: \emph{(i)} closed-ended questionnaires (to capture demographic information such as age, sex, country, etc.), \emph{(ii)} hourly self-reports throughout the day (to capture hourly mood, social context, etc.), and \emph{(iii)} sensor data (continuous sensing modalities such as activity type, step count, Bluetooth, WiFi, location, cellular, and proximity; and interaction sensing modalities including app usage, touch events, screen on/off episodes, and notifications.). All sensor measurements were aggregated with self-reports to create features characterizing the time window during which the report occurred. The final dataset we obtained has over 100 features (for more detailed information on the data processing pipeline and extracted features, please refer to \cite{meegahapola2023generalization}). Regarding missing data in the context of smartphone sensing, it can arise due to various reasons such as the device being in low-consumption mode, sensor failure, user privacy settings, airplane mode, or hardware limitations in certain phone models. To address this issue, feature modalities with more than 70\% missing data, namely Bluetooth low energy, Bluetooth normal, Cellular GSM, and Cellular WCDMA, were dropped similar to prior work \cite{santani2018drinksense, assi2023complex}. More details regarding the used features can be found in Appendix~\ref{appendix:wenet_data}.

\subsection{WEEE: Multimodal Wearable Sensing Dataset for Energy Expenditure Estimation}

The WEEE dataset was collected from 17 participants (12 men and 5 women) by the authors of \cite{gashi2022multidevice}. The processed version of the dataset we used in this paper was obtained from the authors of \cite{amarasinghe2023multimodal}. The data collection process involved capturing information during the execution of three specific activities: resting, cycling, and running. Participants were equipped with eight different wearable devices, including an Indirect Calorimeter device, which served as the ground-truth measurement for energy expenditure estimation. Alongside sensor data, the dataset also encompasses demographic and body composition details, activity specifics, and questionnaire-based data obtained from each participant. Despite the use of eight devices during data collection, only three devices were selected for this study due to inconsistencies, missing data, and also because they contain comparable multimodal data, allowing us to conduct domain adaptation. These devices and their respective sensors are abbreviated as follows: \textit{i)} Nokia Bell Labs Earbuds: accelerometer, gyroscope, PPG; \textit{ii)} Empatica E4 Wristband: accelerometer, PPG; \textit{iii)} Muse S Headband: accelerometer, gyroscope. Hence, the objective is to infer and estimate a person's energy expenditure in a particular moment (the ground truth values come from VO2 Master Analyzer Face Mask: VO2 (ml/kg/min) as the gold standard ground truth\footnote{In machine learning terms, the term "gold-standard" refers to the highest level of accuracy or reliability in ground truth labels or annotations. Gold-standard annotations are often obtained using methods that are considered highly reliable or accurate, such as expert manual annotations, precise measurements, or comprehensive and well-established criteria \cite{wu2023udama, dy2023domain}.}), based on data from wearable data. Source and target domain combinations were determined based on the availability of common sensors in wearable devices. Further, more details regarding the used features can be found in Appendix~\ref{appendix:weee_data}.

\subsection{Domains and Inference Tasks}

Our intention here is to delineate the various experimental settings encompassing multiple datasets and inferences. As depicted in Table~\ref{tab:datasets_source_target}, our focus is on two specific datasets: WENET and WEEE.

\begin{table}[t]
    \centering
    \caption{Summary of datasets, source and target domains, modalities used, and the performed inferences. C stands for classification and R stands for regression.}
    \label{tab:datasets_source_target}
    \resizebox{0.9\textwidth}{!}{%
    \begin{tabular}{lllll}
         \rowcolor[HTML]{EDEDED}
         \textbf{Dataset} &
         \textbf{Source} & 
         \textbf{Targets} &
         \textbf{Modalities} &
         \textbf{Inferences}
         \\

        \arrayrulecolor{Gray}

        WENET &
        Italy & 
        \makecell[l]{India, China, \\Mexico, Paraguay, \\UK, Denmark} & 
        \makecell[l]{Location, Bluetooth, Wifi, \\Cellular, Notifications, Proximity,\\Activity Type, Steps, Screen Events,\\User presence, Touch events, App events} & 
        \makecell[l]{Mood (C) \cite{meegahapola2023generalization}\\Social Context (C) \cite{meegahapola2020alone, kammoun2023understanding, mader2024learning}} \\

        \arrayrulecolor{Gray}
        \midrule
        
        WENET &
        Mongolia & 
        \makecell[l]{India, China,\\Mexico, Paraguay,\\UK, Denmark} & 
        \makecell[l]{Location, Bluetooth, Wifi, \\Cellular, Notifications, Proximity,\\Activity Type, Steps, Screen Events,\\User presence, Touch events, App events} &  
        \makecell[l]{Mood (C) \cite{meegahapola2023generalization}\\Social Context (C) \cite{meegahapola2020alone, kammoun2023understanding, mader2024learning}} \\

        \arrayrulecolor{Gray}
        \midrule

        WEEE &
        EarBuds & 
        Empatica & 
        \makecell[l]{Accelerometer, Photoplethysmography} & 
        \makecell[l]{EEE (R) \cite{amarasinghe2023multimodal}}  \\
        
        \arrayrulecolor{Gray}
        \midrule

        WEEE &
        EarBuds & 
        Muse & 
        \makecell[l]{Accelerometer, Gyroscope} & 
        \makecell[l]{EEE (R) \cite{amarasinghe2023multimodal}} \\
        
        \arrayrulecolor{Gray}
        \bottomrule
        
    \end{tabular}
    }
\end{table}

WENET dataset was employed to facilitate domain transfer across distinct countries, a problem setting motivated by previous studies \cite{meegahapola2023generalization, assi2023complex}. Accordingly, we considered Italy ($N_{italy}$=151,342 from 240 users) and Mongolia ($N_{mongolia}$=94,006 from 214 users) as source domains. This was done because these two countries have larger datasets. Multiple target domains, namely India ($N_{india}$=4,233 from 39 users), China ($N_{china}$=22,289 from 41 users), Mexico ($N_{mexico}$=11,662 from 20 users), Paraguay ($N_{paraguay}$=9,744 from 28 users), UK ($N_{uk}$=26,688 from 72 users), and Denmark ($N_{denmark}$= 10,010 from 24 users) were considered for each source domain. These countries have comparatively smaller datasets compared to sources. This also shows how the technique performs well with limited target data. Consequently, our analysis spanned a total of 12 source-target pairs within the WENET dataset. For each pair, we undertook two classification tasks previously defined: a two-class mood inference (positive vs. negative) \cite{meegahapola2023generalization} and a two-class social context inference (alone vs. with others) \cite{mader2024learning, meegahapola2020alone, kammoun2023understanding}. Both these tasks hold significance in the context of digital health and mobile food diary applications. It is worth noting that the ground truth for the inferences are: i) mood, which is subjective, and silver-standard \footnote{In machine learning terms, the term "silver-standard" is used to describe annotations or ground truth labels that are of lower accuracy or reliability compared to the gold-standard because of uncertainty, bias, and/or noise. They are still considered useful and informative, and are commonly used in inferences. These annotations might be obtained through less stringent methods, such as automated algorithms, self-reports, surrogate measures, or indirect observations \cite{wu2023udama, dy2023domain}.} because it is captured with self-reports; and ii) social context, which is more objective, but still silver-standard because of self-reports.  

Our approach with the WEEE dataset revolved around domain transfer across diverse devices sharing common sensor modalities. This task is also motivated by prior work that highlights the importance of domain adaptation for devices across body positions \cite{chang2020systematic, fortes2022learning}. It is noteworthy that not all devices in the original dataset possess identical sensors. For instance, the Nokia Bell Labs earbuds comprise an accelerometer, gyroscope, and PPG sensor, while the Empatica E4 wristband lacks a gyroscope. Consequently, our experiments encompass two setups: firstly, treating EarBuds ($N_{earbuds}$=9,226 from 17 users) as the source domain and Empatica ($N_{empatica}$=9,226 from 17 users) as the target domain, utilizing accelerometer and PPG data from both devices; secondly, considering EarBuds as the source domain and the Muse S headband ($N_{muse}$=9,226 from 17 users) as the target domain, leveraging accelerometer and gyroscope as the sensing modalities. It is worth noting that data from all devices were collected simultaneously from different body positions, hence the same number of data points. In both setups, we adopted energy expenditure estimation \cite{weee} as the target inference, constituting a regression task. This inference is further characterized and validated in prior studies \cite{amarasinghe2023multimodal, albinali2010using, dannecker2013comparison, o2020well}. It is worth noting that the ground truth here is the gold standard for energy expenditure estimation. 

\section{RQ1: Using Statistical Tests to Quantify Distribution Shift of Sensors}\label{sec:rq1}

\subsection{Methodology}\label{subsec:rq1_methods}

The aim of this analysis is to provide empirical evidence for the rationale behind the development of a multi-branch architecture. In accordance with previous studies \cite{varshney2022trustworthy}, two primary methods for quantifying distribution shift are statistical tests \cite{nanchen2023keep} and inference performance metrics \cite{meegahapola2023generalization}. Statistical test-based techniques are known for their cost-effectiveness and ability to offer a general estimation of the shift for each sensing modality \cite{nanchen2023keep}. Thus, we could employ common statistical tests such as t-test \cite{kim2015t}, PERMANOVA and PERMDISP \cite{nanchen2023keep}, and Cohen's-d \cite{cohen1988statistical} to assess the distribution shift of sensor modalities for each target inference. In this context, after an initial analysis of these tests, we selected Cohen's-d for our analysis due to its relatively linear distribution of values \cite{meegahapola2023generalization, assi2023complex, nanchen2023keep}, within a range approximating 0 and 1. Most importantly, it allowed the best downstream performance for domain adaptation. In addition, the rule of thumb of 0.8 or above: large effect size, 0.5: moderate effect size, and 0.2: small effect size allows easy interpretation \cite{cohen1988statistical}. This characteristic facilitated the utilization of normalized values in our architecture for $\lambda_m$ (Section~\ref{subsec:training_process}). Please note that we use the terms distribution shift or shift between two modalities interchangeably to refer to Cohen's-d from this point onward.

\begin{figure}
    \centering
    
    \begin{minipage}{0.8\textwidth}
        \centering
        \includegraphics[width=\linewidth]{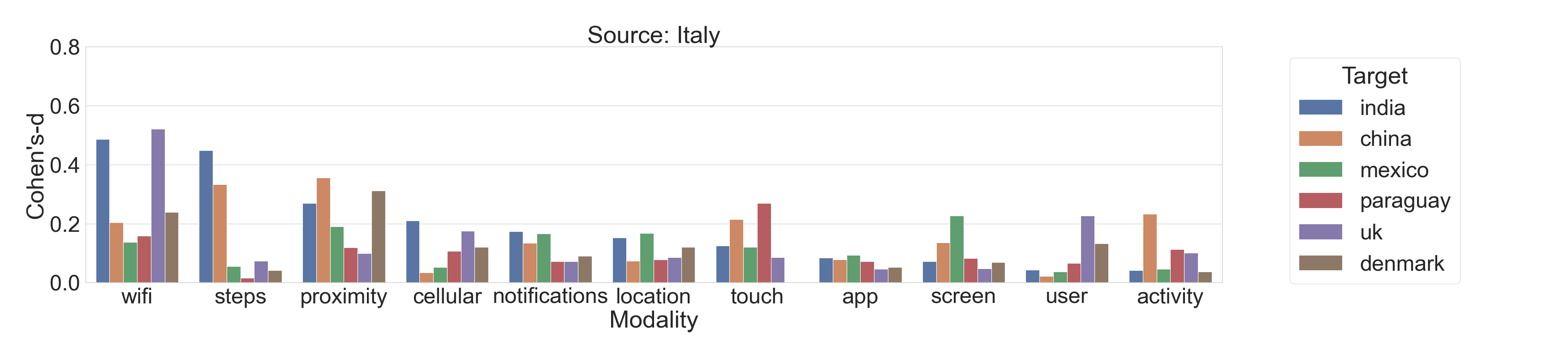}
        \caption{Average Cohen's-d values for modalities. Italy is the source domain.}
        \label{fig:italy_modalities}
    \end{minipage}
    \hfill
    \begin{minipage}{0.8\textwidth}
        \centering
        \includegraphics[width=\linewidth]{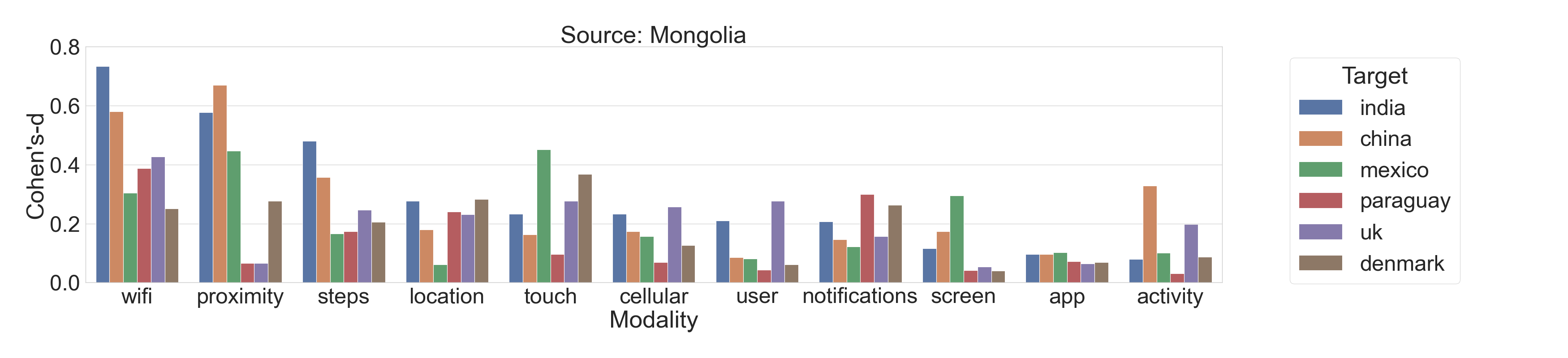}
        \caption{Average Cohen's-d values for modalities. Mongolia is the source domain.}
        \label{fig:mongolia_modalities}
    \end{minipage}
    \hfill
    \begin{minipage}{\textwidth}
        \centering
        \includegraphics[width=0.8\linewidth]{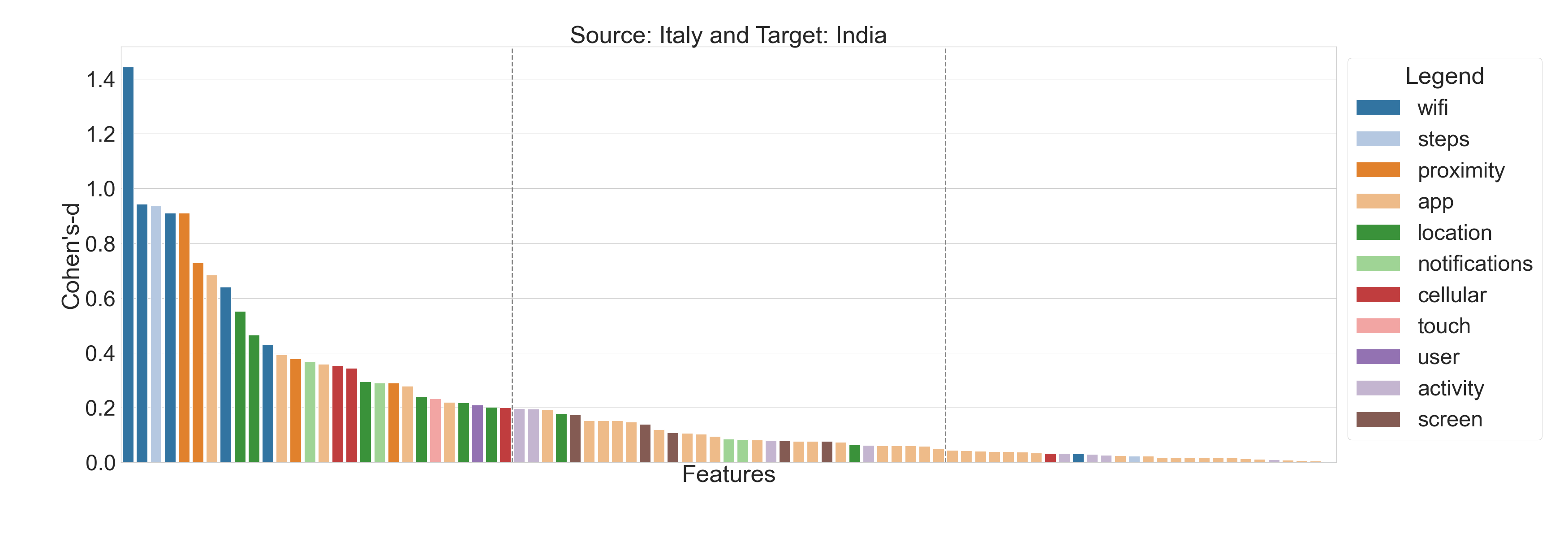}
        \caption{Cohen's-d Values Italy and India, sorted in descending order. Modalities are marked in different colors.}
        \label{fig:italy_india}
    \end{minipage}
\end{figure}

For the WENET dataset, our initial step involved calculating Cohen's-d values for all captured features. Subsequently, we aggregated these values by computing the mean for each modality (e.g., wifi, steps, proximity, location, etc.). By designating Italy and Mongolia as source domains, we plotted the results for other target domains (Figure~\ref{fig:italy_modalities} and Figure~\ref{fig:mongolia_modalities}). This approach enabled us to comprehend how modalities could exhibit varying degrees of distribution shifts for the same source and target domains. The underlying concept is to demonstrate that aggregating features by modalities facilitate the differentiation of high and low levels of distribution shifts. Continuing, we proceeded to visualize Cohen's-d values for all features, assigning distinct modality-specific colors to each bar for enhanced clarity (Figure~\ref{fig:italy_india}). This visualization aimed to provide insights into whether distribution shifts often emanate from the same set of modalities or if there are instances of outliers with high Cohen's-d values from specific modalities exhibiting relatively low levels of distribution shift overall. Due to space limitations, we only show the distribution for the case when Italy is the domain, and India is the target domain. 

For the WEEE dataset, we did a similar analysis. We aggregated Cohen's-d values by computing the mean for each modality (e.g., acc, ppg, gyro --- Figure~\ref{fig:earbuds_modalities}). It is worth noting that, in the setup of transferring from EarBuds to Empatica, only accelerometer and PPG data are available. In the other case of EarBuds to Muse, only accelerometer and gyroscope data are available as common features. Continuing, we proceeded to visualize Cohen's-d values for all features, assigning distinct modality-specific colors to each bar for enhanced clarity. This was done separately for pairs EarBuds and Empatica (Figure~\ref{fig:earbuds_empatica}) and EarBuds and Muse (Figure~\ref{fig:earbuds_muse}). 

\subsection{Results}\label{subsec:rq1_results}

For WENET, Figures~\ref{fig:italy_modalities} and~\ref{fig:mongolia_modalities} present the quantification of distribution shift using Cohen's-d values, which indicate the effect size. The x-axes represent modalities, while the y-axes show the shift. Each modality is color-coded to represent the target domain. Notably, the figures reveal that various countries have diverse modalities that exhibit the highest and lowest distribution shifts compared to Italy. For instance, WiFi is prominent for India and the UK, proximity for China and Denmark, screen for Mexico, and touch for Paraguay. This pattern remains consistent when the source is Mongolia, except for touch in Mexico and Denmark, and WiFi in Paraguay. Moreover, when analyzing shifts within the target countries, the values for different modalities contrast. For instance, with Italy as the source, the target domain India exhibits WiFi and steps with a Cohen's-d of around 0.5 (medium effect size). Conversely, all modalities such as app, screen, user, and activity have values below 0.1 (very small effect size), indicating minimal distribution shift. Similar trends are observed for other countries, even when Mongolia is the source. However, it is important to note that these diagrams do not account for individual features within modalities, which could have high distribution shifts, but their impact might be mitigated by numerous other features within the same modality with low shifts. This aspect is demonstrated in Figure~\ref{fig:italy_india}, where the distribution shift of each feature for source Italy and target India is plotted, with colors denoting modalities. While we can not visualize all such features across various source-target domain pairs, we have identified a considerable number of such cases that highlight significant shifts despite the overall low shift in the modality.

\begin{figure}
    \centering
    \begin{minipage}{0.33\textwidth}
        \centering
        \includegraphics[width=\linewidth]{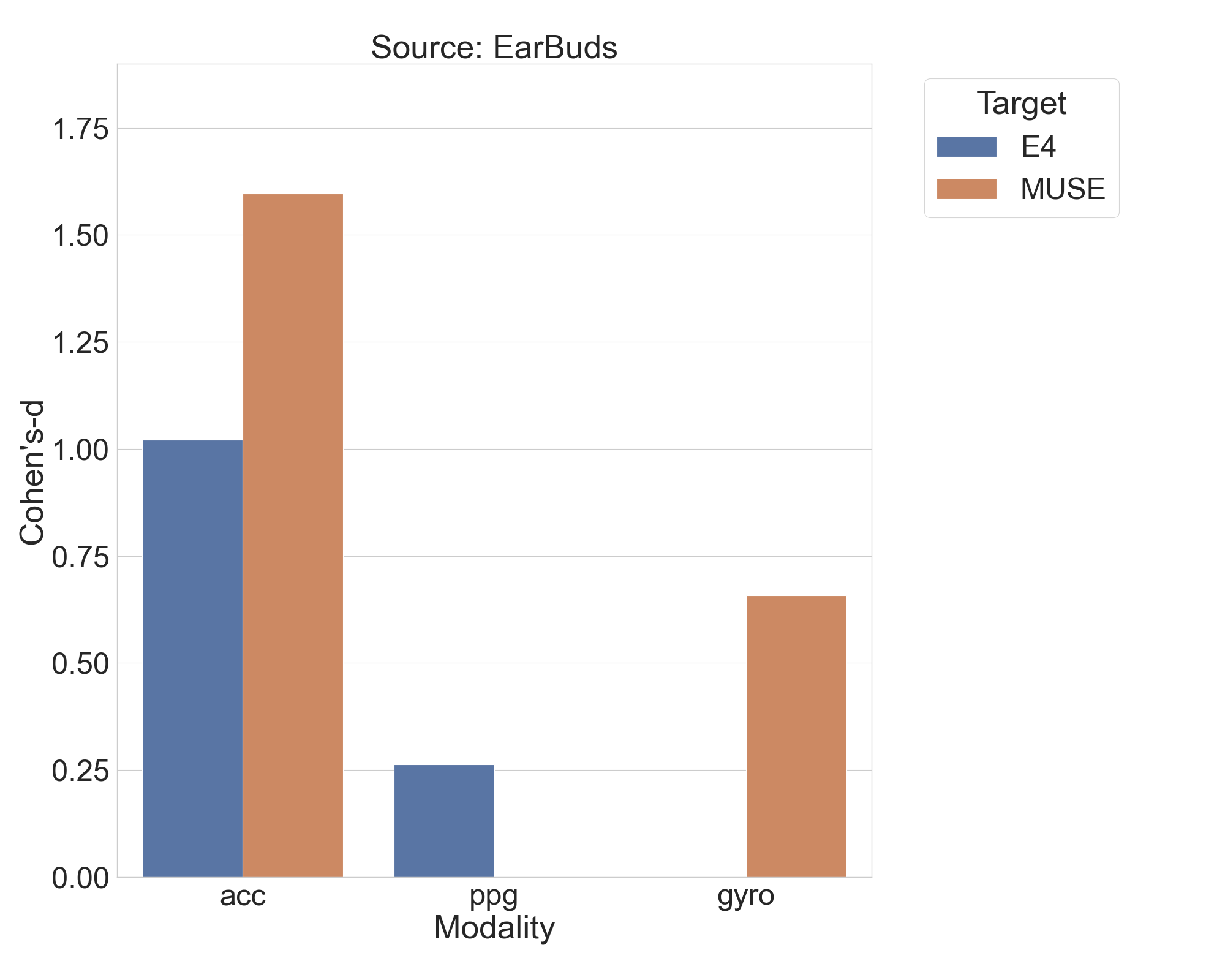}
        \caption{Average Cohen's-d values for modalities. EarBuds is the source domain}
        \label{fig:earbuds_modalities}
    \end{minipage}
    \hfill
    \begin{minipage}{0.3\textwidth}
        \centering
        \includegraphics[width=\linewidth]{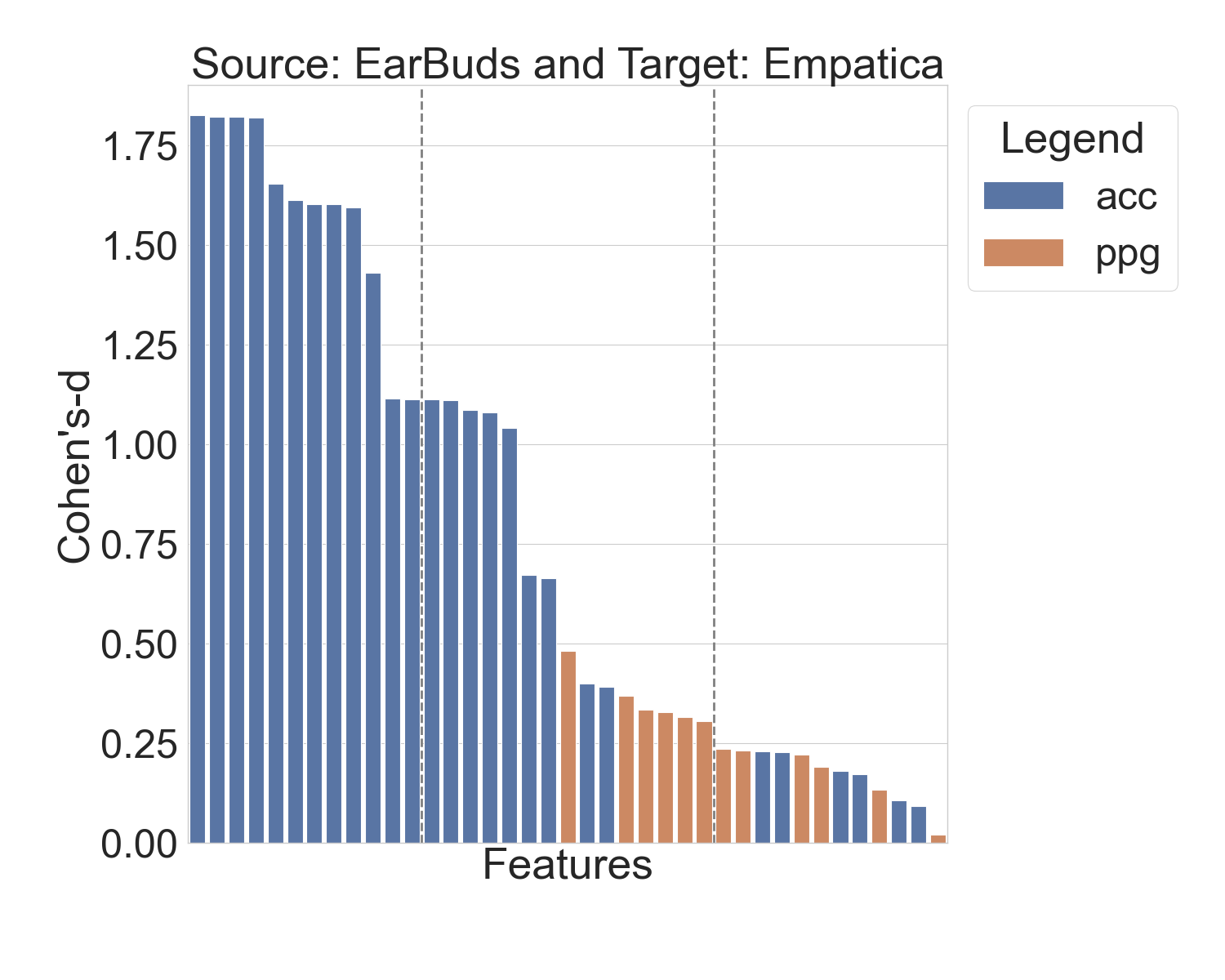}
        \caption{Cohen's-d values for EarBuds and Empatica, sorted in descending order. Modalities are marked in different colors.}
        \label{fig:earbuds_empatica}
    \end{minipage}
    \hfill
    \begin{minipage}{0.33\textwidth}
        \centering
        \includegraphics[width=0.9\linewidth]{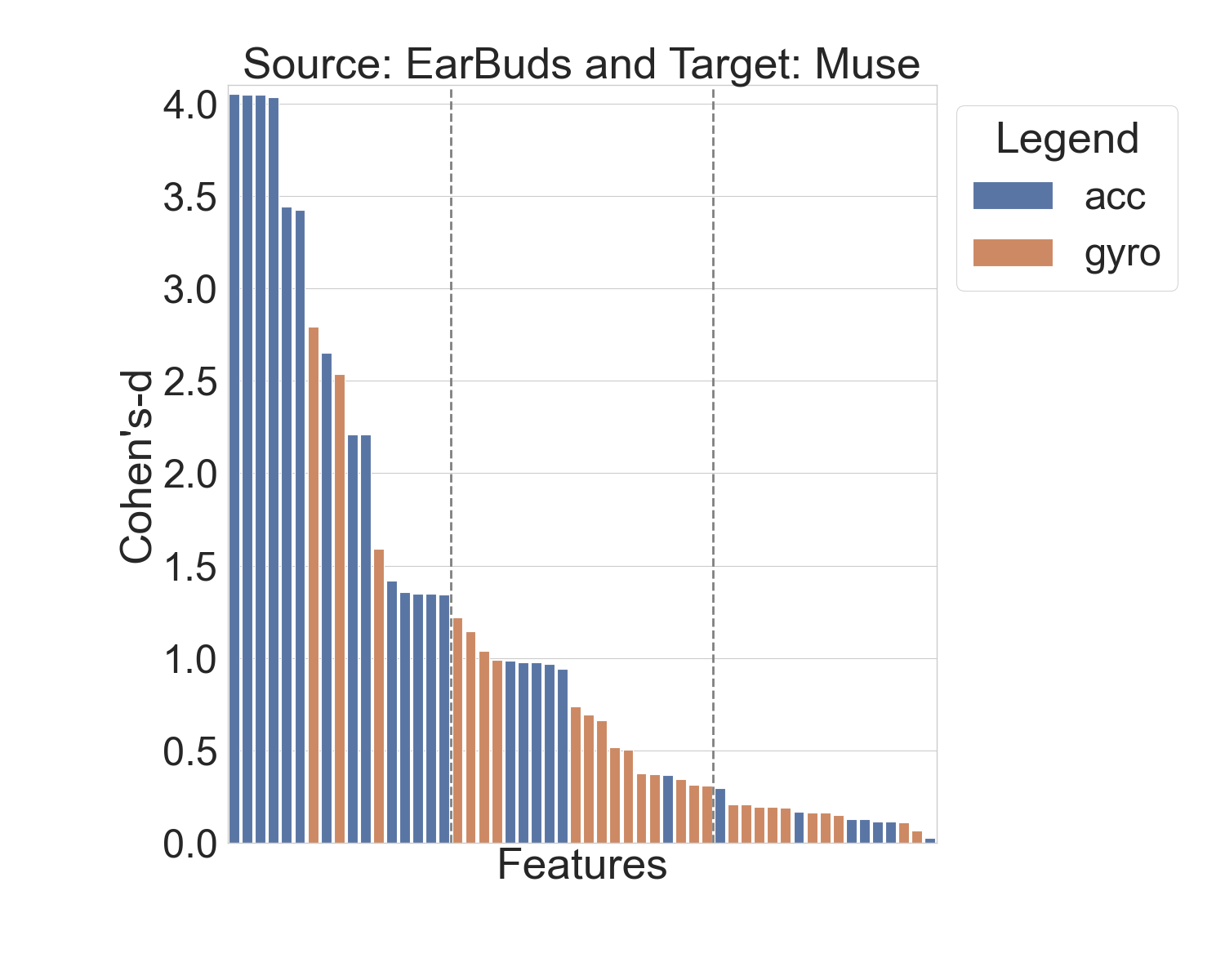}
        \caption{Cohen's-d values for EarBuds and Muse, sorted in the descending order. Modalities are marked in different colors.}
        \label{fig:earbuds_muse}
    \end{minipage}
\end{figure}

Moving to the WEEE dataset, in Figure~\ref{fig:earbuds_modalities}, we illustrate the modality-specific shifts for the source-target pairs of Earbuds and Empatica, as well as Earbuds and Muse. For the EarBuds and Empatica pair, it is evident that accelerometer features exhibit a substantial shift (Cohen's-d of $\approx$1—indicating a large effect size), whereas PPG features demonstrate only a smaller shift (Cohen's-d of $\approx$0.2—reflecting a small effect size). Similarly, a contrasting pattern is observed for the EarBuds and Muse pair, with accelerometer and gyroscope modalities showcasing noticeable shifts. Notably, even the gyroscope exhibits a considerable shift in this context (Cohen's-d around 0.6—indicating an above-medium effect size). To delve deeper into this analysis, we present feature-specific shifts in Figure~\ref{fig:earbuds_empatica} and Figure~\ref{fig:earbuds_muse}. Regarding the EarBuds and Empatica pair, fewer outliers are observed compared to the modality-specific pattern. Most features with above-medium effect sizes primarily originate from the accelerometer, while PPG features tend to exhibit small to medium effect sizes, indicating smaller shifts. Conversely, in the case of the EarBuds and Muse pair, the features exhibit a more mixed distribution, deviating from the modality-specific shifts highlighted in Figure~\ref{fig:earbuds_modalities}. 

In summary, in answering RQ1, the statistical analysis suggests that it might be possible to categorize features into groups with high, medium, or low distribution shifts based on modalities or by considering mixed feature groups derived from multiple modalities with prominent effect sizes. These insights directly address the design of the architecture that we use to answer RQ3, where we propose the use of multiple branches for distinct feature groups, as detailed in Section~\ref{subsec:multibranch}.
\section{RQ2: Domain Adversarial Training with Multimodal Sensing Features}\label{sec:rq2}

\subsection{Methodology}\label{subsec:rq2_methods}

As the next step, we performed domain adversarial training using the base architecture described in Section~\ref{subsec:uda_classification}. Our experiments were implemented in Python, with TensorFlow \cite{tensorflow2015whitepaper}, Keras \cite{chollet2015keras}, and PyTorch \cite{paszke2019pytorch}. The architecture consists of an encoder without considering the multimodality of the data. Our dataset splitting involved separating training (70\%) and testing (30\%) sets to ensure non-overlapping users, facilitating leave-k-out cross-validation \cite{hastie2009elements}. We conducted five such random training and testing splits to ensure robustness and reported the average results. This experimental setup is similar to the approach proposed in \cite{chang2020systematic}, with the distinction that we employed processed tabular features from multiple modalities, similar to \cite{wu2023udama, meegahapola2021one}, instead of the raw sensor values with a feature extractor.

For the WENET dataset, we initiated our analysis by training models on the training sets for Italy and Mongolia as source domains. Our model architecture was designed to infer Mood and Social Context, with intermediate layer sizes of 128, 128, and 64, all with the ReLU activation \cite{agarap2018deep}. Dropout \cite{srivastava2014dropout} was used with rates of 50\%, 50\%, and 20\% to mitigate overfitting. We used sigmoid activation \cite{dubey2022activation} and binary cross-entropy loss function \cite{wang2020comprehensive}, suitable for the two-class nature of our inferences. Adam optimizer \cite{kingma2014adam} and a batch size of 32 was chosen. We also implemented early stopping to prevent overfitting after five epochs of non-improving validation loss within \{0, 300\} epochs. Performance evaluation employed the area under the receiver operating characteristic curve (AUC) with macro averaging, which considers class imbalance \cite{assi2023complex, meegahapola2023generalization}. Evaluation of the models began by assessing the performance of the \textbf{S}ource model on the \textbf{S}ource testing set (\textbf{S->S}). These results were averaged across the five iterations. Subsequently, we evaluated the \textbf{S}ource model on the \textbf{T}arget datasets (\textbf{S->T}). Due to multiple targets for each source, we averaged the results. We also fine-tuned the source model on target training sets with transfer learning and evaluated on the target domain testing set (\textbf{S->T (w/ TL)}). Note that this setup is not unsupervised and needs labels in the target domain. We then proceeded with domain adversarial training (\textbf{DANN} \cite{ganin2016domain}), as outlined in Section~\ref{subsec:training_process}---Step 3a, where we first trained the encoder (with layer sizes 128 and 64) and later the target classifier (with intermediate layer sizes 64 and 32) and domain classifier (with intermediate layer sizes 64 and 32). These classifiers employed sigmoid activation and binary cross-entropy loss at their respective final layers. A fixed $\lambda=1$ was used for gradient reversal during encoder training. Data with labels from the source domain contributed to the loss for both domain and target classifiers during training, while unlabelled target domain data only contributed to the domain classifier loss. 

Even for WEEE dataset, the methodology was similar to that of the WENET dataset, albeit with smaller models due to the dataset's size. We trained models for EarBuds, utilizing accelerometer and PPG data, and for EarBuds with accelerometer and gyroscope data. These modality combinations were chosen to allow domain adaptation for two devices, as described in Table~\ref{tab:datasets_source_target}. The models were designed to infer energy expenditure estimation, with intermediate layer sizes of 64 and 32, and using the ReLU activation function. Dropout with rates of 30\% and 20\% was used for regularization. The mean squared error loss function was utilized given that it is a regression—Adam optimizer and a batch size of 16 facilitated model training. Early stopping was also implemented. The evaluation process mirrored that of the WENET dataset. \textbf{DANN}, following the process outlined in Section~\ref{subsec:training_process}, involved training the encoder (with layer sizes 64 and 32) and later the target regressor (with intermediate layer size 32) and domain classifier (with intermediate layer sizes 32 and 16). The target regressor employed mean squared error, and the domain classifier employed sigmoid activation and binary cross-entropy loss at their final layers. 

Hence, in summary, the inferences that we conducted across both datasets are given below. 
\begin{itemize}
    \item \textbf{S->S}: performance of the model trained in the source domain, for the source testing set. This provides an upper bound for the possible results in the target domain.
    \item \textbf{S->T (w/ TL)}: performance of the model trained in the source domain for the target testing set after fine-tuning the target training set with transfer learning. This setup assumes that labels are available in the target domain, hence could lead to higher performance and act as another ceiling for performance. It is also worth noting that ground truth labels used in training models can be gold-standard or silver-standard as mentioned in Section~\ref{sec:datasets}. In WENET, ground truth is the silver standard as they are self-reports that can be noisy. However, mood reports can be subjective and social context reports are more objective, potentially leading to differences in the accuracy and noisyness \cite{meegahapola2020smartphone}. In WEEE, the ground truth is the gold standard. 
    \item \textbf{Random}: performance of a random classifier/regressor on the target testing set. 
    \item \textbf{S->T}: performance of the model trained in the source domain for the target testing set. 
    \item \textbf{DANN \cite{ganin2016domain}, MMD \cite{chang2020systematic}, ADDA \cite{tzeng2017adversarial}}: performance of the model trained in the source domain and unlabelled target data (training set), on the target testing set. Methods used include domain adversarial training (DANN) as suggested in earlier sections and \cite{ganin2016domain}; maximum mean discrepancy (MMD) loss-based adaptation as suggested by \cite{chang2020systematic}; and adversarial discriminative domain adaptation (ADDA) as suggested by \cite{tzeng2017adversarial}. These techniques have been shown to perform reasonably in sensing-based inference tasks \cite{wu2023udama, pillai2024investigating, chang2020systematic}.  
\end{itemize}

In addition to the aforementioned results, we have also included the state-of-the-art performances for \textbf{S->S} and \textbf{S->T}, as reported in the original papers that proposed these specific tasks, using the same datasets. Mood inference results were extracted from \cite{meegahapola2023generalization}, social context inference from \cite{mader2024learning}, and the EEE from \cite{amarasinghe2023multimodal}. It is important to note that the original study conducted experiments using only five countries for social context inference. Consequently, the \textbf{S->T} performance reported therein cannot be directly compared with the performance presented here. Furthermore, all the aforementioned papers achieved the best-performing models using classic machine learning techniques, such as random forest classifiers and XGBoost models. While these models yield high performance on the datasets, they cannot be directly applied to the domain adaptation algorithms under investigation in this work.

\subsection{Results}\label{subsec:rq2_results}

\begin{table}[t]
    \centering
    \caption{Results for classification tasks. Results are presented as average AUC scores (higher the better). TL refers to transfer learning where labelled target domain data are available.}
    \label{tab:results_classification}
    \resizebox{0.9\textwidth}{!}{%
    \begin{tabular}{llllllll}
        \arrayrulecolor{Gray}
        \toprule
        \textbf{Dataset} & 
        \textbf{WENET} &
        \textbf{WENET} &
        \textbf{WENET} &
        \textbf{WENET} \\
        
        \textbf{Inference} & Mood & Mood & Social Context & Social Context \\
        \textbf{Source} & Italy & Mongolia & Italy & Mongolia \\
        \textbf{Targets} & Avg. of Countries & Avg. of Countries & Avg. of Countries & Avg. of Countries \\

        \rowcolor[HTML]{EDEDED}
        \multicolumn{5}{c}{\textbf{RQ2}} \\
        
        \textbf{Random}                       &
        0.45$\pm$0.12 \textcolor{orange!70}{\rule{0.06cm}{0.2cm}} &
        0.46$\pm$0.08 \textcolor{orange!70}{\rule{0.08cm}{0.2cm}} &
        0.42$\pm$0.14 \textcolor{orange!70}{\rule{0.00cm}{0.2cm}} &
        0.44$\pm$0.11 \textcolor{orange!70}{\rule{0.02cm}{0.2cm}} \\

        \hline 
        
        \textbf{S->S}                         &
        0.61$\pm$0.04 \textcolor{orange!50}{\rule{0.38cm}{0.2cm}} &
        0.59$\pm$0.06 \textcolor{orange!50}{\rule{0.34cm}{0.2cm}} &
        0.63$\pm$0.03 \textcolor{orange!50}{\rule{0.46cm}{0.2cm}} &
        0.65$\pm$0.04 \textcolor{orange!50}{\rule{0.46cm}{0.2cm}} \\

        \textbf{S->S -- Meegahapola et al. \cite{meegahapola2023generalization}}                         &
        0.55$\pm$0.05 \textcolor{orange!50}{\rule{0.26cm}{0.2cm}} &
        0.49$\pm$0.08 \textcolor{orange!50}{\rule{0.17cm}{0.2cm}} &
        - &
        -
        \\

        \textbf{S->S -- Mader et al. \cite{mader2024learning}} &
        - &
        - &
        0.66$\pm$0.02 \textcolor{orange!50}{\rule{0.48cm}{0.2cm}} &
        0.73$\pm$0.06 \textcolor{orange!50}{\rule{0.52cm}{0.2cm}} \\

        \hline
        
        \textbf{S->T}                         &
        0.46$\pm$0.08 \textcolor{orange!75}{\rule{0.08cm}{0.2cm}} &
        0.48$\pm$0.07 \textcolor{orange!75}{\rule{0.16cm}{0.2cm}} &
        0.44$\pm$0.03 \textcolor{orange!75}{\rule{0.02cm}{0.2cm}} &
        0.49$\pm$0.08 \textcolor{orange!75}{\rule{0.18cm}{0.2cm}} \\
        
        \textbf{S->T (w/ TL)}                 &
        0.51$\pm$0.05 \textcolor{orange!60}{\rule{0.18cm}{0.2cm}} &
        0.52$\pm$0.06 \textcolor{orange!60}{\rule{0.20cm}{0.2cm}} &
        0.56$\pm$0.05 \textcolor{orange!60}{\rule{0.28cm}{0.2cm}} &
        0.55$\pm$0.07 \textcolor{orange!60}{\rule{0.26cm}{0.2cm}} \\

        \textbf{S->T -- Meegahapola et al. \cite{meegahapola2023generalization}}                         &
        0.48$\pm$0.01 \textcolor{orange!50}{\rule{0.16cm}{0.2cm}} &
        0.50$\pm$0.00 \textcolor{orange!50}{\rule{0.17cm}{0.2cm}} &
        - &
        -
        \\

        \textbf{S->T -- Mader et al. \cite{mader2024learning}} &
        - &
        - &
        0.57$\pm$0.06 \textcolor{orange!50}{\rule{0.29cm}{0.2cm}} &
        0.55$\pm$0.08 \textcolor{orange!50}{\rule{0.26cm}{0.2cm}} \\

        \hline

        \textbf{MMD \cite{chang2020systematic}}  &
        0.49$\pm$0.08 \textcolor{orange!95}{\rule{0.20cm}{0.2cm}} &
        0.54$\pm$0.08 \textcolor{orange!95}{\rule{0.22cm}{0.2cm}} &
        0.51$\pm$0.04 \textcolor{orange!95}{\rule{0.20cm}{0.2cm}} &
        0.52$\pm$0.09 \textcolor{orange!95}{\rule{0.28cm}{0.2cm}} \\

        \textbf{ADDA \cite{tzeng2017adversarial}}  &
        0.49$\pm$0.11 \textcolor{orange!95}{\rule{0.20cm}{0.2cm}} &
        0.50$\pm$0.07 \textcolor{orange!95}{\rule{0.22cm}{0.2cm}} &
        0.51$\pm$0.08 \textcolor{orange!95}{\rule{0.20cm}{0.2cm}} &
        0.51$\pm$0.06 \textcolor{orange!95}{\rule{0.28cm}{0.2cm}} \\

        \textbf{DANN \cite{ganin2016domain}}  &
        0.52$\pm$0.07 \textcolor{orange!95}{\rule{0.20cm}{0.2cm}} &
        0.53$\pm$0.02 \textcolor{orange!95}{\rule{0.22cm}{0.2cm}} &
        0.52$\pm$0.03 \textcolor{orange!95}{\rule{0.20cm}{0.2cm}} &
        0.54$\pm$0.05 \textcolor{orange!95}{\rule{0.28cm}{0.2cm}} \\

        \rowcolor[HTML]{EDEDED}
        \multicolumn{5}{c}{\textbf{RQ3}} \\
        \textbf{Ours ($\lambda=1$, Setup1)}               &
        0.55$\pm$0.05 \textcolor{orange!50}{\rule{0.26cm}{0.2cm}} &
        0.52$\pm$0.06 \textcolor{orange!50}{\rule{0.24cm}{0.2cm}} &
        0.55$\pm$0.06 \textcolor{orange!50}{\rule{0.26cm}{0.2cm}} &
        0.57$\pm$0.05 \textcolor{orange!50}{\rule{0.30cm}{0.2cm}} \\
        
        \textbf{Ours (w/ $\lambda_m$, Setup1)}    &
        0.56$\pm$0.04 \textcolor{orange!60}{\rule{0.28cm}{0.2cm}} &
        0.53$\pm$0.07 \textcolor{orange!60}{\rule{0.26cm}{0.2cm}} &
        0.56$\pm$0.04 \textcolor{orange!60}{\rule{0.28cm}{0.2cm}} &
        0.57$\pm$0.05 \textcolor{orange!60}{\rule{0.30cm}{0.2cm}} \\
        
        \textbf{Ours (w/ $\lambda_m$, Setup2)}    &
        0.58$\pm$0.04 \textcolor{orange!90}{\rule{0.32cm}{0.2cm}} &
        0.54$\pm$0.03 \textcolor{orange!90}{\rule{0.28cm}{0.2cm}} &
        0.55$\pm$0.05 \textcolor{orange!90}{\rule{0.26cm}{0.2cm}} &
        0.55$\pm$0.03 \textcolor{orange!90}{\rule{0.26cm}{0.2cm}} \\
        
        \arrayrulecolor{Gray}
        \bottomrule
    \end{tabular}
    }
\end{table}

Table~\ref{tab:results_classification} presents the classification outcomes for the WENET dataset, for mood and social context inferences. The results show the model's performance under different scenarios. Specifically, \textbf{S->S} showcases how the model performs when evaluated on the source testing set. The inferred values fall within the range of 0.59 to 0.61 AUC. Although the performance in the source domain is not notably high, this aligns with trends observed in recent research focused on mental well-being and contextual inference using multimodal mobile sensing datasets \cite{xu2023globem, meegahapola2023generalization}. Despite not achieving high levels of performance, these results still provide a foundation for investigating domain adaptation techniques, where even small enhancements in performance on target domains are crucial. \textbf{S->T} is where the model trained on the source domain is evaluated on the target domain's testing set. As expected, performance experiences a decline across all four inferences compared to \textbf{S->S}. With transfer learning fine-tuning (\textbf{S->T (w/ TL)}), the performance improves across all four inferences as expected because it uses labels in the target domain. Interestingly, the application of \textbf{DANN} leads to further performance enhancements across social context inferences while showing a slight performance decline compared to \textbf{S->T (w/ TL)} for mood inference. This could be because mood labels are more subjective; hence even having labels in the target domain is less useful, whereas \textbf{DANN} leads to marginally better results. However, for social context, which is more objective ground truth, having labels led to increased performance, even more than the adapted model with \textbf{DANN}. Hence, while further work is needed, this could suggest that the label source quality and objectivity might have an effect on fine-tuning or domain adaptation performance. 

Table~\ref{tab:results_regression} presents the regression outcomes for the WEEE dataset, focusing on energy expenditure estimation (EEE) inference. In this context, EarBuds serves as the source domain, while Empatica or Muse serves as the target domain. The source domain performance (\textbf{S->S}) yields MAE values of 0.62 and 0.59 for EarBuds, representing the desired ceiling performance. The random baseline, on the other hand, delivers poor results. In the \textbf{S->T} scenario, the performance exhibits a reduction of approximately 0.17 MAE and 0.21 MAE for Empatica and Muse, respectively. Notably, the application of transfer learning (\textbf{S->T (w/ TL)}) results in improved performance compared to \textbf{DANN}. This stands in contrast to mood inference in the WENET dataset results, where \textbf{DANN} marginally outperformed transfer learning for mood inference. The divergence in results can be attributed to the nature of labels; the WENET dataset employs silver standard labels derived from user self-reports, however, with high and low subjectivity for mood and social context, respectively, while both labels could also be susceptible to noise. On the other hand, the WEEE dataset leverages gold-standard labels derived from lab-based measurements. Our findings underscore that while transfer learning with silver standard labels, especially when having subjective ground truth, does not universally guarantee performance gains, it could potentially provide an alternative ceiling performance (a baseline for the maximum performance the model could reach). It is noteworthy, however, that transfer learning necessitates the availability of labels in the target domain, which is not the primary scenario we are addressing.

In conclusion, when tackling RQ2, our exploration revealed that domain adversarial training applied to multimodal mobile sensing datasets translates to enhanced performance compared to \textbf{S->T} setting. Notably, for mood inference in the WENET dataset, the observed increase even surpassed/equaled that achieved through transfer learning. This phenomenon can likely be attributed to the presence of silver standard, and potentially subjective labels in this dataset. On the other hand, when examining the WEEE dataset, \textbf{DANN} contributed to improved performance, although not reaching the same level as transfer learning, which was to be expected. This discrepancy could be attributed to the presence of high-quality gold standard labels available in both the source and target domains, which were effectively utilized for training models. This calls for further research in this direction in the future. 
\section{RQ3: Multi-Branch Domain Adversarial Training}\label{sec:rq3}

RQ3 in our study is motivated by the nuanced findings from both RQ1 and RQ2, which highlight the variable degrees of domain shift across different modalities. These insights suggest that distinctively treating modalities or feature groups could yield improved performance. This hypothesis stems from the understanding that modalities with lesser distribution shifts may benefit from minimal adaptation, while those with higher shifts might require more intensive adaptation strategies. A single encoder structure, as used in DANN, does not allow for this level of tailored adaptation across modalities. Therefore, RQ3 explores the concept of multiple encoder branches, each adjusted for varying degrees of shifts, to address these disparities. This approach aims to optimize the adaptation process by treating each modality according to its specific domain shift characteristics, potentially leading to more effective overall performance.

\subsection{Methodology}\label{subsec:rq3_methods}

\subsubsection{Experiments with Multiple Branches}\label{subsubsec:rq3_methods_1}

The next step is to replace the encoder with multiple branches as described in Section~\ref{subsec:multibranch}. We considered experimental approaches with two distinct setups. These setups were based on the results we obtained for experiments in Section~\ref{sec:rq1}. There, we discussed how distribution shift could be quantified by aggregating features based on modalities (e.g., activity, steps, wifi, location, etc.) or by aggregating based on feature level shift as quantified by statistical tests (e.g., top 33\% of features, bottom 33\% of features, and the rest, etc. regardless of the modality). 

\textit{Setup1---Branches Based on Modalities:} In the WENET dataset, there are over ten modalities. Having separate branches for all modalities leads to a complex optimization problem. Hence, in this paper, we focus on having three branches for which we were able to obtain decent results. Beyond the three branches, we did not obtain good results, as it became a difficult optimization given the dataset size and the challenging nature of the task, as described in Section~\ref{sec:rq2}. Hence, when having three branches, for each target country, we used the modality with the highest shift as one branch, the modality with the lowest shift as another branch, and the rest of the modalities in one branch, as visualized in Figure~\ref{fig:archi1}. When considering modalities, we normalized the highest Cohen's-d modalities $\lambda_m$ to 1, and if the Cohen's-d of the lowest modality was below 0.2 (below small effect size), normalized it such that it is $\lambda_m=0$ (all source-target pairs had below 0.1 modalities, as shown in Figure~\ref{fig:italy_modalities} and Figure~\ref{fig:mongolia_modalities}). $\lambda_m$ for the set of features in the middle was normalized to a suitable value, as described in Section~\ref{subsec:training_process}---Step 3c. We only have two modalities in the WEEE dataset for both inferences. Hence, we used two branches and again normalized between 0 and 1 to obtain the $\lambda_m$ values for the two modalities, following Section~\ref{subsec:training_process}. Finally, for this setup, we first conducted experiments with $\lambda=1$ for all branches (\textbf{$\lambda=1$, Setup1}). Then, we used different $\lambda_m$ values for branches based on the shift and conducted experiments (\textbf{w/ $\lambda_m$, Setup1}).

\begin{wraptable}{r}{0.6\textwidth}
    \centering
    \caption{Results for regression tasks. Results are presented as mean absolute errors (MAE) (the lower the better).}
    \label{tab:results_regression}
    \resizebox{0.6\textwidth}{!}{%
    \begin{tabular}{llllllll}
        \arrayrulecolor{gray}
        \toprule
        \textbf{Dataset} & 
        \textbf{WEEE} &
        \textbf{WEEE} 
        \\
        
        \textbf{Inference} &
        EEE &
        EEE \\
        
        \textbf{Source} &
        EarBuds &
        EarBuds \\
        
        \textbf{Target} &
        Empatica &
        Muse \\

        \rowcolor[HTML]{EDEDED}
        \multicolumn{3}{c}{\textbf{RQ2}} \\

        \textbf{Random}                      &
        1.35$\pm$0.31 \textcolor{blue!70}{\rule{0.85cm}{0.2cm}} &
        1.41$\pm$0.43 \textcolor{blue!70}{\rule{0.91cm}{0.2cm}} \\

        \arrayrulecolor{gray}
        \hline 
        
        \textbf{S->S}                        &
        0.62$\pm$0.11 \textcolor{blue!30}{\rule{0.12cm}{0.2cm}} &
        0.52$\pm$0.06 \textcolor{blue!30}{\rule{0.12cm}{0.2cm}} \\

        \textbf{S->S -- Amarasinghe et al. \cite{amarasinghe2023multimodal}}                        &
        - &
        0.61$\pm$N/A \textcolor{blue!30}{\rule{0.12cm}{0.2cm}} \\

        \hline 
        
        \textbf{S->T}                        &
        0.79$\pm$0.15 \textcolor{blue!90}{\rule{0.29cm}{0.2cm}} &
        0.73$\pm$0.10 \textcolor{blue!90}{\rule{0.23cm}{0.2cm}} \\
        
        \textbf{S->T (w/ TL)}                &
        0.67$\pm$0.07 \textcolor{blue!50}{\rule{0.17cm}{0.2cm}} &
        0.56$\pm$0.04 \textcolor{blue!50}{\rule{0.06cm}{0.2cm}} \\

        \textbf{S->T -- Amarasinghe et al. \cite{amarasinghe2023multimodal}}                        &
        - &
        -
        \\

        \hline 
        
        \textbf{MMD \cite{chang2020systematic}} &
        0.71$\pm$0.05 \textcolor{blue!90}{\rule{0.17cm}{0.2cm}} &
        0.69$\pm$0.06 \textcolor{blue!90}{\rule{0.16cm}{0.2cm}} \\

        \textbf{ADDA \cite{tzeng2017adversarial}} &
        0.76$\pm$0.05 \textcolor{blue!90}{\rule{0.23cm}{0.2cm}} &
        0.70$\pm$0.16 \textcolor{blue!90}{\rule{0.17cm}{0.2cm}} \\
        
        \textbf{DANN \cite{ganin2016domain}} &
        0.73$\pm$0.05 \textcolor{blue!90}{\rule{0.23cm}{0.2cm}} &
        0.65$\pm$0.06 \textcolor{blue!90}{\rule{0.15cm}{0.2cm}} \\

        \rowcolor[HTML]{EDEDED}
        \multicolumn{3}{c}{\textbf{RQ3}} \\
        \textbf{Ours ($\lambda=1$, Setup1)}              &
        0.74$\pm$0.04 \textcolor{blue!50}{\rule{0.24cm}{0.2cm}} &
        0.62$\pm$0.06 \textcolor{blue!30}{\rule{0.12cm}{0.2cm}} \\
        
        \textbf{Ours (w/ $\lambda_m$, Setup1)}   &
        0.69$\pm$0.06 \textcolor{blue!70}{\rule{0.19cm}{0.2cm}} &
        0.60$\pm$0.03 \textcolor{blue!50}{\rule{0.10cm}{0.2cm}} \\
        
        \textbf{Ours (w/ $\lambda_m$, Setup2)}   &
        0.69$\pm$0.05 \textcolor{blue!70}{\rule{0.19cm}{0.2cm}} &
        0.64$\pm$0.03 \textcolor{blue!70}{\rule{0.14cm}{0.2cm}} \\
        
        \arrayrulecolor{gray}
        \bottomrule
    \end{tabular}
    }
\end{wraptable}

\textit{Setup2---Branches Based on Feature Group:} In both datasets, we could sort all features in descending order based on shift for each source-target pair. Then, we could consider three groups similar to Setup 1, by considering 33\% of data with the highest shift, 33\% of data with the lowest shift, and finally, the rest of the 33\% of data in the middle. While the percentage could be changed, we did not delve deeper into that in this analysis and focused on obtaining equal splits for the three feature sets. In Figures~\ref{fig:italy_india}, \ref{fig:earbuds_empatica}, and \ref{fig:earbuds_muse}, these splits are marked with vertical dotted lines. Finally, for this setup, we used different $\lambda_m$ values for branches, based on the shift, as suggested in Section~\ref{subsec:training_process} (\textbf{w/ $\lambda_m$, Setup2}).

\subsubsection{Experiments with Multiple Modalities in High and Low Branches for Setup1}\label{subsubsec:rq3_methods_2} An identified limitation of Setup1 is evident in instances where a single modality, representing branches with the highest or lowest shift, encompasses only a limited number of features. For instance, when the source domain is Italy and the target is India, the modality with the highest shift, `wifi' ($\lambda_0 = 1$), includes merely seven features. In contrast, the `activity' modality with the lowest shift contains eight features ($\lambda_2 = 0$), while the remaining features (over 80) fall within the moderate shift range ($\lambda_1 = 0.62$). This scenario results in an imbalance among the branches. However, such an imbalance does not manifest in Setup2, where the equitability of sizes across branches is maintained. We performed another experiment to assess the potential implications of this limitation on performance. Specifically, we introduced additional modalities to the high and low-shift branches. To accomplish this, we define $\alpha$ to indicate the number of modalities present within the high and low shift branches. Thus, $\alpha=1$ corresponds to one modality each in the high and low branches, $\alpha=2$ signifies two modalities each, $\alpha=3$ represents three modalities each within these branches, and so on. Subsequently, we conduct a series of experiments similar to those described in Section~\ref{subsubsec:rq3_methods_1}, systematically varying the $\alpha$ values. This experimentation enables us to gauge the influence of modifying the number of modalities on performance. Given the inadequacy of modalities within the WEEE dataset, it is important to underscore that this specific experiment pertains only to the WENET dataset.

\subsection{Results}\label{subsec:rq3_results}

\begin{wrapfigure}{r}{0.4\textwidth}
  \begin{center}
    \centering
    \includegraphics[width=\linewidth]{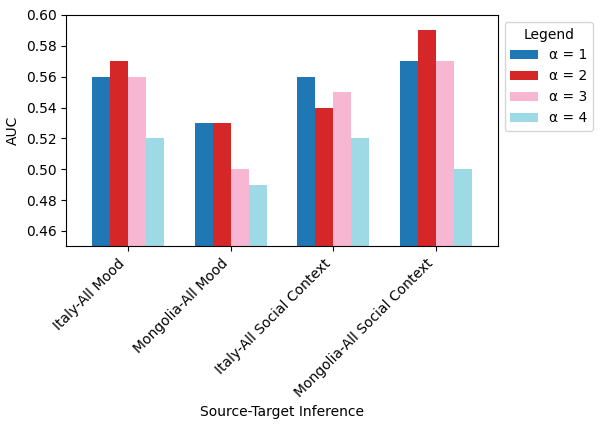}
    \caption{Inference results for various $\alpha$ values.}
    \label{fig:alpha_changes}
  \end{center}
\end{wrapfigure}

\subsubsection{Experiments with Multiple Branches}\label{subsubsec:rq3_results_1} The results for the WENET dataset are presented in Table~\ref{tab:results_classification}. When $\lambda$ is set to 1, indicating domain adversarial training with uniform $\lambda$ values across all branches, the performance exceeds that of \textbf{DANN} in all instances except for mood inference with Mongolia as the source. Within Setup1, the introduction of distinct $\lambda_m$ values for branches, determined by modality distribution, yields minor performance enhancements across all scenarios compared to the uniform $\lambda=1$ configuration, except for social context inference with Mongolia as the source where both setups yielded an AUC of 0.57. Moreover, the superiority of Setup1 over Setup2 is not clear, as each configuration displayed better performance in different inferences. However, a discernible trend emerged with mood inference, which is a more subjective task reliant on nuanced labeling, indicating that fine-tuning with labels (\textbf{S->T (w/ TL)}) failed to elevate performance, even relative to \textbf{DANN} and our proposed approach. Conversely, transfer learning achieved performance on par with our method for the objective ground truth of social context inference, again underscoring the possible impact of ground truth nature on inference accuracy as discussed in Section~\ref{subsec:rq2_results}. From another sense, this highlights that our technique achieves decent performance, even compared to transfer learning, for social context inferences in the WENET dataset. 

The findings for the WEEE dataset are shown in Table~\ref{tab:results_regression}. Here, our approach once again outperformed \textbf{DANN} and other baselines across both inferences. However, the performance does not surpass that of fine-tuning with transfer learning. This discrepancy can be attributed to the presence of gold-standard ground truth labels in this dataset, allowing fine-tuning to exhibit superior performance. Furthermore, distinguishing between the efficacy of Setup1 and Setup2 remains inconclusive, despite Setup1's superior performance for the Earbuds and Muse source-target pair. As mirrored in the results for the WENET dataset, even here, employing diverse $\lambda$ values for the branches proved to be more effective than adopting a uniform $\lambda=1$ strategy. Thus, the adjustment of $\lambda$ based on branch-specific shift statistics yielded more better results.

\subsubsection{Experiments with Multiple Modalities in High and Low Branches for Setup1}\label{subsubsec:rq3_results_2} 

The outcomes of the experiment conducted to explore the effects of varying $\alpha$ values are presented in Figure~\ref{fig:alpha_changes}. The results exhibit distinct patterns across different inferences and source-target domain pairings. In specific scenarios, such as the 'Italy-All Mood' and 'Mongolia-All Social Context', employing $\alpha=2$ yielded slightly superior results compared to when $\alpha$ was set to 1. Similarly, for the 'Mongolia-All Mood' inference, $\alpha=2$ yielded results similar to those of $\alpha=1$, while $\alpha=3$ led to notably improved performance. Intriguingly, in all instances, setting $\alpha=4$ resulted in consistently subpar performance. These findings underscore the nuanced nature of determining optimal configurations for the number of branches and the number of modalities within each branch during model training. There appears to be no universal formula for these selections, as their efficacy depends on the specific inference and source-target domains. Nevertheless, a recurring trend across the analyses is that employing multiple branches with an appropriately chosen $\alpha$ and $\lambda$ consistently outperforms utilizing a single encoder with fixed $\lambda$, except in very few experiments (e.g., WEEE Earbuds-Empatica DANN=0.73 performed better than $\lambda=1$, Setup1; WEEE Earbuds-Muse $\lambda=1$, Setup1 performed better than w/ $\lambda_m$ Setup2; WENET Mood w/ source Mongolia DANN = 0.53 performed similar to $\lambda=1$, Setup1 and w/ $\lambda_m$, Setup1).

As a summary, in answering RQ3, the analysis shed light on the nuanced interplay between diverse factors such as the nature of ground truth (gold/silver standard, subjective/objective), modality distributions and related distribution shifts, and branch-specific adaptation dynamics. Hence, the conducted experiments provide evidence that incorporating multiple branches for different feature sets, based on either modality or distribution shift-based feature groups, yields improved performance in unsupervised domain adaptation. The results for both the WENET and WEEE datasets consistently indicate that this approach outperforms baseline methods and enhances domain adaptation performance. Further, the findings emphasize the importance of adaptability in adjusting parameters such as $\lambda$ and $\alpha$ based on specific contexts, inference tasks, and source-target domain pairings. This adaptability proves crucial in effectively harnessing the advantages of domain adversarial training, highlighting its potential to significantly enhance model generalization and performance across diverse multimodal mobile sensing datasets.

\section{Discussion}\label{sec:discussion}

\subsection{Implications}

The use of multi-branch encoders with varying lambda values in domain adaptation carries both modeling and practical implications. First, we discuss a few modeling implications related to our proposed approach.  

\begin{enumerate}[wide, labelwidth=!, labelindent=0pt]
    \item Adaptation Strategy Customization: The insights drawn from the experiments underscore the importance of adopting good adaptation strategies. The observed performance improvements achieved through tuning parameters such as $\lambda$ and $\alpha$ values emphasize the importance of tailoring the training process to the specific characteristics of the data and the nature of the inference tasks. The results highlight the need for flexibility in adaptation techniques, acknowledging that different data modalities and inference tasks may demand distinct strategies for optimal performance. This implication encourages researchers and practitioners to consider the intricacies of the data and the problem domain when crafting adaptation methodologies, contributing to a more effective approach to domain adaptation. Moreover, in some of our experiments that were not reported in the study, we modified the annealing schedule of $\lambda$ suggested by Ganin et al. \cite{ganin2016domain, ganin2015unsupervised}, by both doubling and halving its rate to evaluate the impact on model performance. This adjustment aimed to discern whether variations in the rate at which \(\lambda\) transitions from its initial to final value would affect the training dynamics or the model's final evaluation metrics, such as AUC in classification or MAE in regression. Our findings revealed that altering the annealing schedule did not yield results with substantial or clear conclusions. Specifically, doubling the annealing rate resulted in marginally lower training durations without commensurate benefits in model performance. However, the observed differences in AUC and MAE were negligible, suggesting that while the rate of annealing influences training time, it did not impact the model's effectiveness according to these metrics. This could be a matter of further exploration in the future. Finally, it is also worth noting that introducing separate parameters across branches, as discussed in Section~\ref{sec:rq3}, led only to a minor increase in performance for both regression and classification tasks. This observation prompts the question of whether the added complexity of this setup justifies its implementation. We argue that it does, particularly considering that the maximum performance enhancement observed for the WENET dataset, transitioning from S->S to S->T, was approximately 15\% (See Table~\ref{tab:results_classification}). In this context, even a marginal improvement of 1\% is reasonable. Future research could further investigate these findings across other datasets and tasks that may present less of a challenge, to determine this approach's broader applicability and benefit. 
    \item Impact of Ground Truth Nature: The divergent performance trends observed across various inference tasks, particularly distinguishing between subjective tasks like mood inference, more objective tasks like social context inference, and gold standard ground truth-based tasks like energy expenditure estimation, provide evidence that confirms the impact of ground truth nature on the domain adaptation model's performance. These findings highlight the intricate interplay between ground truth labels' quality and reliability and domain adaptation methods' success. The outcomes showcase the need to take into account the inherent subjectivity and potential noise in ground truth labels, particularly in contexts where human judgment and perception play a crucial role. The implications underscore the pivotal role of domain adaptation not only in mitigating distribution shifts but also in accommodating the peculiarities of the ground truth data. 
\end{enumerate}

In addition, the following practical implications could be considered in the deployment of this kind of model. 

\begin{enumerate}[wide, labelwidth=!, labelindent=0pt]
    \item Enhancing Real-world Applicability: The performance improvements obtained by the multi-branch adaptation strategies have implications for real-world applications reliant on multimodal mobile sensing data. These findings suggest that practitioners integrating data from diverse sensors could effectively enhance the utility of their models by customizing their adaptation strategies based on the nuanced characteristics of the data and the demands of specific inference tasks. This adaptability offers a concrete means to increase predictive accuracy in data-driven applications such as mental well-being monitoring, personalized healthcare, and behavior analysis. By fine-tuning adaptation techniques to the intricacies of data distributions and domain shifts, practitioners can achieve more robust and meaningful outcomes in these critical domains.
    \item Guidance for Model Design: The insights obtained from the experiments offer valuable guidance for practitioners and researchers engaged in designing and deploying adaptation models for multimodal data. By comprehending the impact of factors such as ground truth quality and modality distribution on model performance, informed decisions could be made about the architecture and configuration of their models. This understanding streamlines experimentation, minimizes trial-and-error efforts, and accelerates the development of effective adaptation techniques tailored to specific use cases. 
\end{enumerate}

In conclusion, the theoretical and practical implications derived from the experimental results collectively underscore the inherent flexibility, adaptability, and performance-enhancing capabilities of the M3BAT architecture for unsupervised domain adaptation. These implications advance the understanding of adaptation techniques and provide actionable insights for researchers seeking to use these strategies to improve model performance and generalization capabilities. As the field of multimodal mobile sensing continues to mature, the insights derived from these implications inform the development of sophisticated techniques that effectively address the complexities of domain shifts in real-world settings.

\subsection{Limitations and Future Work}

In this paper, we have chosen to focus on DANN as our primary technique for multimodal mobile sensing, a decision guided by its proven effectiveness in addressing domain shifts and enhancing dataset generalization. DANN, as an adversarial-based method, is particularly suited for the variability and complexity inherent in mobile sensing data. While domain adversarial training is a foundational approach, it is important to note that it may not fully encapsulate the subtleties of real-world domain variations. This limitation, however, does not undermine the value of DANN in our study; rather, it highlights the necessity of exploring a range of domain adaptation techniques, both discrepancy-based and adversarial-based, as suggested by Goel et al. \cite{goel2023unsupervised} and Wu et al. \cite{wu2023udama}. Our choice of DANN is motivated by its balance between theoretical robustness and practical applicability, making it a fitting starting point for our exploration into domain adaptation in mobile sensing. Nonetheless, we acknowledge the importance of further research into other methods. Techniques stemming from adversarial discriminative domain adaptation \cite{tzeng2017adversarial}, self-ensembling \cite{perone2019unsupervised}, and moment matching \cite{peng2019moment} represent potential avenues for future investigation. These methods could offer alternative or complementary strategies for more intricate and challenging domain shifts. By focusing on DANN, we believe that this study sets a foundational groundwork for such explorations, emphasizing the need to continually evolve and enhance domain adaptation methodologies.

The reliance on prior probability shift or label shift assumptions, though insightful, may not fully capture the complexities of real-world data distributions. This challenge is not exclusive to ubiquitous and mobile sensing---however, the unique data characteristics and complexities in this field underscore the need to investigate alternative assumptions or methodologies in situations with ambiguous label distributions. This approach is critical to more effectively address the diverse range of domain variations and challenges. Recent studies in ubicomp have pointed out the necessity for such research \cite{nanchen2023keep}. It is also important to note that the techniques we have proposed in this paper are not specifically designed to handle label shifts comprehensively, although they partially address them. Therefore, future research should delve deeper into these aspects to enhance our understanding and management of label shifts in these contexts. Further, the current methodology leverages separate branches for individual sensor modalities (Setup1), a pragmatic approach that effectively considers the unique characteristics of each modality. However, exploiting potential synergies between modalities remains a compelling avenue for future research. Even though this was partially done with Setup2, the development of more sophisticated multimodal fusion techniques could enable a more comprehensive integration of information across different sensor sources, potentially yielding further performance gains. Scalability and applicability in real-world deployment scenarios are two considerations for the practical utility of the technique. To this end, future research should prioritize refining the method’s efficiency and robustness in diverse real-world settings. 

Another limitation of our current approach in domain adversarial training is the requirement for a large, well-labeled dataset from the source domain. This necessity arises because the effectiveness of domain adversarial training hinges significantly on the model's ability to learn comprehensive and generalized feature representations from the source domain. These representations are crucial as they enable the model to perform well on the target domain, which might differ substantially in terms of data distribution. A large dataset in the source domain ensures a diverse range of examples, encompassing a wide variety of features and scenarios. This diversity is key to training a robust model that can effectively reduce the domain discrepancy by aligning the feature distribution of the source domain with that of the target domain. In our case, we specifically chose Italy and Mongolia in the WENET datasets as our source datasets due to their extensive and diverse data pools. These datasets provide a rich set of features and examples that aid in the development of a model capable of generalizing well to other countries. Conversely, smaller datasets might not offer the same level of diversity and comprehensiveness, potentially leading to a model that is less effective at generalizing across domains. This limitation is particularly pronounced in domain adversarial training, where the goal is to learn from the source domain and to bridge the gap to a potentially different target domain.

One more limitation is that we exclusively used Cohen's-d to quantify distribution shifts. However, this choice was made with careful consideration. We selected Cohen's-d as our primary statistical method because all the data we used, such as activity and app usage, are numeric (detailed in the Appendix). Cohen's-d is effective for this type of data. For the sensor data from the accelerometer and gyroscope in the WEEE dataset, we calculated statistical features that reflect specific characteristics within time windows and then applied Cohen's-d to these features. In the case of the periodic PPG signal, we focused on extracting numeric features such as blood volume pulse and heart rate. This makes Cohen's-d a relevant and suitable choice for our analysis. It is also important to note that Cohen's-d, with values typically taking a range closer to 0-1, could be seen as a valuable metric for approximately quantifying distribution shift, given that $\lambda$ value also falls in the same range. However, Cohen's-d is not the only option that could be used for this purpose. In our previous research \cite{nanchen2023keep}, we considered other statistical measures like the t-statistic, permanova, and permadisp as alternatives for quantifying distribution shifts. \lakmal{Moreover, while we have clarified the scope of our analysis to focus on numerical statistical features across both datasets, it is crucial to emphasize that more techniques beyond Cohen's-d should be considered when there are categorical input features. Therefore, future studies might benefit from exploring statistical measures beyond Cohen's-d, for such features. Hence, even though intuitive, the current mapping between Cohen's-d and $\lambda$ is naive and rule-based, as described in Section~\ref{subsubsec:cohensd_to_lambda}. This could also be improved in future studies.}

Expanding the technique's versatility and usefulness could involve extending it to encompass transfer learning and personalized model scenarios. This broader scope could cater to diverse application needs and user-specific requirements, enhancing the technique's adaptability and applicability. For example, given our findings, sometimes it might make sense to do domain adaptation first and personalize a model to target users in a target domain rather than directly personalizing a model. These directions need further investigation. Additionally, the technique's application has been primarily centered around time series data processed through time windows, as is customary in this context across many studies \cite{wu2023udama, meegahapola2023generalization, spathis2019passive, assi2023complex}. Future research could explore its adaptability when confronted with raw time series data alongside convolutional neural network-based feature extraction, similar to how multimodal data are handled in \cite{xu2023globem}. This expansion would offer insights into the method's effectiveness under different data representations and processing techniques. Furthermore, while the technique has been demonstrated within the domain of mobile sensing, it could also be applicable to tabular datasets with multimodal attributes. Future work could explore its utility across other domains. In conclusion, while the proposed domain adaptation technique exhibits promising results, these limitations and future research lines provide impetus to advance multimodal domain adaptation methodologies in the context of mobile sensing.

\section{Conclusion}\label{sec:conclusion}

In this work, we have explored the effectiveness of a multi-branch domain adaptation technique for multimodal mobile sensing data. Our experiments on the WENET and WEEE datasets highlight the adaptability of the approach through parameter customization, leading to enhanced performance and generalization. The results underscore the need for tailored adaptation strategies, while the distinction between subjective and objective tasks emphasizes the role of ground truth quality. The technique's potential for scenarios with limited labeled data and its applicability to practical settings further demonstrate its significance. However, challenges remain, and future research should focus on refining the technique's scalability, real-world deployment, and fusion of multimodal data. As mobile sensing gains momentum in various domains, this study contributes to the advancement of unsupervised domain adaptation with M3BAT architecture, with a range of additional, real-world machine learning applications.

\begin{acks}

This work was funded by the European Union’s Horizon 2020 WeNet project, under grant agreement 823783. We thank all the volunteers across the world for their participation in the project.

\end{acks}

\bibliographystyle{ACM-Reference-Format}
\bibliography{citations}

\appendix

\newpage

\section{WENET Dataset Features \cite{meegahapola2023generalization}}\label{appendix:wenet_data}

\begin{table}[h]
    \caption{Summary of features extracted from sensing data, aggregated around activity self-reports using a time window.}
    \label{tab:agg-features}
    \begin{tabular}{{p{0.139\linewidth}p{0.725\linewidth}}}
        
        \cellcolor[HTML]{EDEDED}\textbf{Modality} &
        \cellcolor[HTML]{EDEDED} \textbf{Description}
        \\

        \makecell[l]{Location} & 
        \makecell[l]{radius of gyration, distance traveled, mean altitude} \\
\arrayrulecolor{Gray}
    \midrule
    
        \makecell[l]{Bluetooth \\{[}low energy, \\normal{]}} &
        \makecell[l]{number of devices (the total number of unique devices found), mean/std/min/max \\rssi (Received Signal Strength Indication -- measures how close/distant other \\devices are)} \\

\arrayrulecolor{Gray}
    \midrule
    
        \makecell[l]{WiFi} & 
        \makecell[l]{connected to a network indicator, number of devices (the total number of unique \\devices found), mean/std/min/max rssi} \\
        
\arrayrulecolor{Gray}
    \midrule
    
        \makecell[l]{Cellular {[}GSM, \\WCDMA, LTE{]}} & 
        \makecell[l]{number of devices (the total number of unique devices found), mean/std/min/max \\phone signal strength} \\
        
\arrayrulecolor{Gray}
    \midrule
    
        \makecell[l]{Notifications} & 
        \makecell[l]{notifications posted (the number of notifications that came to the phone), \\notifications removed (the number of notifications that were removed by the \\user) -- these features were calculated with and without duplicates.} \\
        
\arrayrulecolor{Gray}
    \midrule
    
        \makecell[l]{Proximity} & 
        \makecell[l]{mean/std/min/max of proximity values} \\
        
\arrayrulecolor{Gray}
    \midrule
        \makecell[l]{Activity} & 
        \makecell[l]{time spent doing activities: still, in\_vehicle, on\_bicycle, on\_foot, running, tilting, \\walking, other (derived using the Google activity recognition API \cite{GoogleActivity2022})} \\
        
\arrayrulecolor{Gray}
    \midrule
        \makecell[l]{Steps} & 
        \makecell[l]{steps counter (steps derived using the total steps since the last phone turned on \\at 10 samples per second), steps detected (steps derived using event triggered for \\each new step captured on change)} \\
        
\arrayrulecolor{Gray}
    \midrule
        \makecell[l]{Screen events} &
        \makecell[l]{number of episodes (episode is from turning the screen of the phone on until the \\screen is turned off), mean/min/max/std episode time (a time window could have \\multiple episodes), total time (total screen on time within the time window)} \\
        
\arrayrulecolor{Gray}
    \midrule
        \makecell[l]{User presence} &
        \makecell[l]{time the user is present using the phone (derived using android API that indicate \\whether a person is using the phone or not)} \\
        
\arrayrulecolor{Gray}
    \midrule
        \makecell[l]{Touch events} &
        \makecell[l]{touch events (number of phone touch events)} \\
        
\arrayrulecolor{Gray}
    \midrule
        \makecell[l]{App events} & 
        \makecell[l]{time spent on apps of each category derived from Google Play Store \cite{likamwa2013moodscope, santani2018drinksense}: \\action, adventure, arcade, art \& design, auto \& vehicles, beauty, board, books \& \\
        reference, business, card, casino, casual, comics, communication,
        dating,\\education, entertainment, finance, food \& drink, health \&
        fitness, house, lifestyle,\\maps \& navigation, medical, music, news \&
        magazine, parenting, personalization, \\photography, productivity,
        puzzle, racing, role playing, shopping, simulation, \\social, sports,
        strategy, tools, travel, trivia, video players \& editors, weather, word, \\ not\_found}   \\

        \hline
    \end{tabular}
\end{table}

\newpage 

\section{WEEE Dataset Features \cite{amarasinghe2023multimodal}}\label{appendix:weee_data}

\begin{table}[h]
    \centering
    \caption{Summary of the features used in the analysis.}
    \label{tab:sensor_preprocessing}
    \begin{tabular}{p{3.2cm} l}
         \rowcolor[HTML]{EDEDED}
         \textbf{Modality} &
         \textbf{Description}
         \\

        \arrayrulecolor{Gray}

        Accelerometer &
        \makecell[l]{Statistical features calculated using tsfresh \cite{tsfresh}: sum\_values, median, mean, length, \\standard\_deviation, variance, root\_mean\_square, maximum, absolute\_maximum, minimum}
        \\

        \arrayrulecolor{Gray}
        \hline

        Gyroscope &
        \makecell[l]{Statistical features calculated using tsfresh \cite{tsfresh}: sum\_values, median, mean, length, \\standard\_deviation, variance, root\_mean\_square, maximum, absolute\_maximum, minimum}
        \\

        \arrayrulecolor{Gray}
        \hline
        
        Photoplethysmography &
        \makecell[l]{Features derived using HeartPy \cite{heartpy}: bpm, ibi, sdnn, sdsd, rmssd, pnn20, \\pnn50, hr\_mad, sd1, sd2, s, sd1/sd2, breathingrate}
        \\

        \arrayrulecolor{Gray}
        \hline

        \arrayrulecolor{Gray}
        \hline
    \end{tabular}
\end{table}

\end{document}